\documentclass[letterpaper, 10 pt, conference]{ieeeconf}  %

\IEEEoverridecommandlockouts                              %

\usepackage{graphics} % for pdf, bitmapped graphics files
\usepackage{graphicx}
\usepackage{graphbox}
\usepackage{amsmath} % assumes amsmath package installed
\usepackage{amssymb}  % assumes amsmath package installed
\usepackage{color}
\usepackage{mathtools}
\usepackage{bm}
\usepackage{algorithm}
\usepackage{algpseudocode}
\usepackage{multirow}
\usepackage{booktabs}
\usepackage{dsfont}
\usepackage{array}
\usepackage{url}

\newcolumntype{P}[1]{>{\raggedright\arraybackslash}p{#1}}

\newcommand{\R}{\mathbb{R}}
\newcommand{\D}{\mathcal{D}}

\newcommand{\x}{\bm{x}} % input
\newcommand{\s}{\bm{s}} % state
\newcommand{\y}{\bm{y}} % output
\newcommand{\w}{\bm{\theta}} % weights
\newcommand{\ac}{\bm{u}} % actions
\newcommand{\Feat}{\Phi} % hidden state
\newcommand{\ep}{\bm{\varepsilon}} % hidden state
\newcommand{\p}{\bm{w}} % hidden state
\newcommand{\Sep}{\Sigma_{\varepsilon}} % hidden state
\newcommand{\pep}{\bm{\nu}} % hidden state
\newcommand{\Spep}{\Sigma_{\nu}} % hidden state

\newcommand{\N}{\mathcal{N}}
\renewcommand{\baselinestretch}{0.96}
\makeatletter
\let\NAT@parse\undefined
\makeatother
\usepackage{hyperref}
\usepackage[capitalise]{cleveref}

\title{\LARGE \bf
Expanding the Deployment Envelope of Behavior Prediction\\via Adaptive Meta-Learning
}

\author{Boris Ivanovic$^{1}$ \hspace{1cm} James Harrison$^{2}$ \hspace{1cm} Marco Pavone$^{1,3}$%
\thanks{$^{1}$Boris Ivanovic is with NVIDIA Research.
        {\tt\small bivanovic@nvidia.com}}%
\thanks{$^{2}$James Harrison is with Google Research, Brain Team.
        {\tt\small jamesharrison@google.com}}%
\thanks{$^{3}$Marco Pavone is with the Department of Aeronautics and Astronautics, Stanford University, and with NVIDIA Research.
        {\tt\small \{pavone@stanford.edu,mpavone@nvidia.com\}}}%
}

\begin{document}

\maketitle
\thispagestyle{empty}
\pagestyle{empty}

\begin{abstract}

Learning-based behavior prediction methods are increasingly being deployed in real-world autonomous systems, e.g., in fleets of self-driving vehicles, which are beginning to commercially operate in major cities across the world.
Despite their advancements, however, the vast majority of prediction systems are specialized to a set of well-explored geographic regions or operational design domains, complicating deployment to additional cities, countries, or continents. 
Towards this end,
we present a novel method for efficiently adapting behavior prediction models to 
new environments.
Our approach leverages recent advances in meta-learning, specifically Bayesian regression, to augment existing behavior prediction models with an adaptive layer that enables efficient domain transfer via offline fine-tuning, online adaptation, or both. %
Experiments across multiple real-world datasets demonstrate that our method can efficiently adapt to a variety of unseen environments.

\end{abstract}

\section{Introduction}

\begin{figure}[t]
  \centering
  \includegraphics[width=1.0\linewidth]{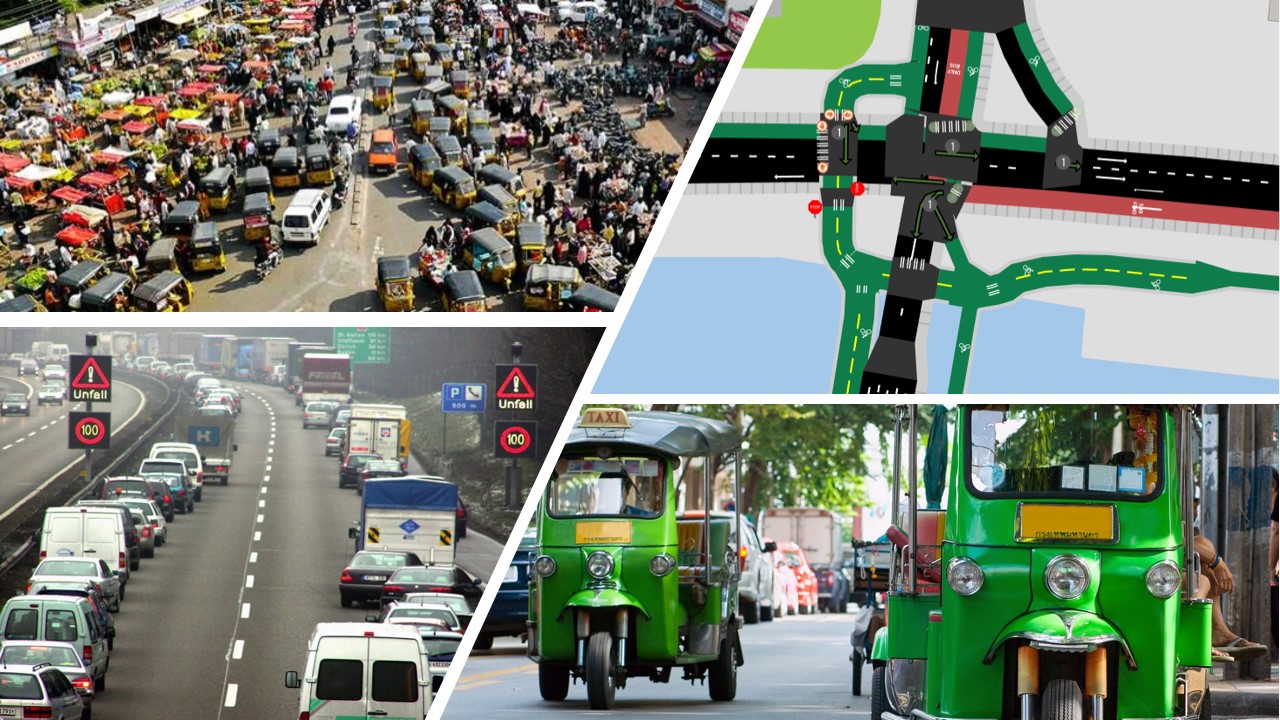}
  
  \vspace{-0.2cm}
  
  \caption{Deploying autonomous systems to a wide variety of diverse locations remains challenging, complicated by (clockwise from top-left) differing social interaction standards (e.g., traffic density in India), environment geometries (e.g., road/bike lane intersections in Ireland), agent types (e.g., Tuk-tuks in Thailand), and laws (e.g., forming a ``Rettungsgasse" during heavy traffic in Germany), among other factors.}
  \label{fig:hero}
  \vspace{-0.5cm}
\end{figure}

Learning-based behavior prediction methods are becoming a staple of the modern robotic autonomy stack, with nearly every
major autonomous vehicle organization incorporating state-of-the-art behavior prediction models in their software stacks~
\cite{GMSafety2018,UberATGSafety2020,LyftSafety2020,ArgoSafety2021,MotionalSafety2021,ZooxSafety2021,NVIDIASafety2021}. In order to deploy such systems safely and reliably alongside humans, organizations extensively train and test models in their specific operational domains. While this improves system accuracy and safety within the targeted domain, it does so at the cost of generalization. Specifically, deploying autonomous vehicles to different cities or countries remains challenging due to their different social interaction standards, laws, agent types, and road geometries (\cref{fig:hero}).

Currently, autonomous vehicles expand their operational domain through an extensive process of collecting and annotating large amounts of data from target locations~\cite{CruiseExpansion2022}. However, manually annotating data is labor-intensive and expensive, and would be prohibitive for a goal of worldwide deployment. To enable widespread deployment without continual data collection and annotation, self-driving vehicles will need to be able to adapt to their surroundings, either from data observed online or from a small amount of already-annotated data.

Towards this end, in this work we present a method for adaptive trajectory forecasting that augments existing prediction models and enables them to efficiently adapt to new environments. Our approach employs meta-learning~\cite{HospedalesAntoniouEtAl2022}, specifically we extend the Bayesian regression approach of ALPaCA~\cite{HarrisonSharmaEtAl2018}, to replace the last layer of a neural network with an adaptive Bayesian linear regression layer whose posterior distribution can be updated efficiently from newly-observed datapoints. As shown in \cref{sec:expts}, our method yields significant error reductions in the face of domain shifts compared to na\"{i}ve transfer and adapts more quickly and efficiently than gradient descent-based fine-tuning.

{\bf Contributions.} Our key contributions are threefold. First, we present a trajectory forecasting algorithm capable of online updating and effective transfer through an adaptive last layer. We combine ALPaCA-based \cite{HarrisonSharmaEtAl2018} last layer adaptation with recurrent models, and show that they work synergistically to enable both broad generalization and efficient transfer (\cref{sec:method}). Second, we present a novel trajectory forecasting algorithm based on Trajectron++~\cite{SalzmannIvanovicEtAl2020} that naturally pairs with the ALPaCA adaptive last layer. In particular, we introduce a within-episode aleatoric uncertainty prediction scheme that enables state-dependent multimodality and modulates last layer adaptation (\cref{sec:multistep_pred}). Finally, we show experimentally that this architecture broadly extends the deployment envelope of trajectory forecasting algorithms, enabling efficient transfer to problem settings substantially different from those on which they were trained (\cref{sec:expts}).

\section{Related Work}

\subsection{Adaptive Behavior Prediction}

The field of behavior prediction has grown significantly in recent years, with many works focusing on improving model accuracy, incorporating social factors (e.g., inter-agent interactions), accounting for multimodality, and leveraging advancements in deep learning methodology~\cite{RudenkoPalmieriEtAl2019}. Current state-of-the-art approaches particularly reflect this, employing graph-structured Recurrent Neural Networks (RNNs) with multimodality modeled by Conditional Variational Autoencoders (CVAEs)~\cite{SohnLeeEtAl2015} (e.g.,~\cite{SalzmannIvanovicEtAl2020}) or Generative Adversarial Nets~\cite{GoodfellowPouget-AbadieEtAl2014} (e.g.,~\cite{KosarajuSadeghianEtAl2019}), Transformers~\cite{VaswaniShazeerEtAl2017} (e.g.,~\cite{GiuliariHasanEtAl2020,YuanWengEtAl2021}), and deep convolutional networks (popular in end-to-end approaches, e.g.,~\cite{ZengLuoEtAl2019}, and for real-time use~\cite{KamenevWangEtAl2022}). While there have been many advances, such works have only been evaluated in the same environments as they were trained.

At the same time, there have been a plethora of datasets released, with thousands of hours of data now publicly available. Accordingly, there have recently been studies into the cross-domain generalization of behavior prediction methods~\cite{GillesSabatiniEtAl2022c}, as well as works tackling the problem of domain adaptation for trajectory forecasting. Broadly, adaptive behavior prediction methods can be categorized into the following three groups:

{\bf Memory.} Memory-based approaches generally tackle domain shifts by first computing past and future trajectory embeddings (e.g., with recurrent neural networks) and then leveraging an associative external memory to store and retrieve the embeddings at test-time~\cite{MarchettiBecattiniEtAl2020,XuMaoEtAl2022}. Trajectory prediction is performed by decoding future embeddings (stored in memory) conditioned on the observed past and any context. Using memory also allows for continuous improvement online by storing novel trajectories. One key drawback of memory-based approaches, especially for online adaptation, is that constantly-growing extra memory
may not be feasible for robotic platforms which have fixed memory and compute limits. In contrast, our method does not require any extra memory.

{\bf Architecture.} Architectural methods generally employ neural networks whose intermediate representations or structures are generalizable and can transfer to different domains. Two recent examples include the Transferable Graph Neural Network (T-GNN)~\cite{XuWangEtAl2022} and the Hierarchical, Adaptable, and Transferable Network (HATN)~\cite{WangHuEtAl2022}. At a high level, both works propose a hierarchical structure, where one component learns domain-invariant trajectory features and another learns low-level, transferable features.
A key drawback of such approaches is that they either require labeled data from the target domain during training-time~\cite{XuWangEtAl2022} or are not inherently online-adaptive~\cite{WangHuEtAl2022}.

{\bf Least Squares.} Methods in this category typically employ recursive least squares to adapt models to data observed online. Similar to our approach, RLS-PAA~\cite{SongZhaoEtAl2011,GoodwinSin2014} updates the last layer of a trained neural network via iterative least squares. However, since RLS-PAA is not performed during training, it is generally only useful for adapting to local agent behaviors within the same dataset (rather than for domain transfer). This is also reflected in RLS-PAA's use and performance in recent literature, e.g.,~\cite{ChengZhaoEtAl2019,SiWeiEtAl2019,AbuduweiliLiu2020} only adapt to local dynamics and agent behaviors within the same dataset. Further, while HATN~\cite{WangHuEtAl2022} tackles cross-domain transfer, its ablation study shows that incorporating RLS-PAA provides no benefit.
In contrast, our method backpropagates \emph{through} the least squares updates during training, yielding a calibrated prior model that can be efficiently adapted to different domains. %

\subsection{Few-Shot Learning by Meta-Learning}

In few-shot learning, a learner must learn to predict from a small amount of data \cite{wang2020generalizing}.
Meta-learning exists as a particular approach to few-shot learning, in which transfer to a new domain is enabled by ``learning to learn" across a set of training tasks \cite{FinnAbbeelEtAl2017,snell2017prototypical, HospedalesAntoniouEtAl2021,duan2016rl,edwards2016towards,ravi2016optimization,santoro2016meta,wang2016learning}. By learning update rules on many tasks, an agent may learn to learn effectively, and thus efficiently, in a new task. 

While there are several ways to design meta-learning algorithms, two are of note for this work. First, \emph{black-box} meta-learning exploits sequence-processing neural network models such as recurrent networks \cite{duan2016rl,wang2016learning,hochreiter2001learning}. While expressive, these models have no particular inductive bias toward learning, and thus they are practically inefficient and require huge amounts of training data. To address this issue, \emph{optimization-based} meta-learners~\cite{FinnAbbeelEtAl2017, HarrisonSharmaEtAl2018, nichol2018first, ravi2016optimization}, which directly leverage optimization in the inner learning loop, have been a major research focus. The model that we develop in this paper can be viewed as a combined black-box and optimization meta-learner. In particular, it was previously shown
that recurrence alone is not sufficient for effective transfer. We address this in this work by showing that recurrence-based adaptation can be effectively paired with optimization-based meta-learning. 

\section{Background: Adaptive Meta-Learning}
\label{sec:bg_adaptive}

Our framework leverages an \textit{adaptive} formulation of meta-learning. In contrast to the more common episodic meta-learning (e.g., MAML~\cite{FinnAbbeelEtAl2017}), the adaptive formulation allows tasks to vary over time within an episode, similar to meta-learning for continual learning~\cite{HarrisonSharmaEtAl2020}. In particular, our approach builds on an extension to ALPaCA~\cite{HarrisonSharmaEtAl2018} developed in~\cite{Harrison2021}. Instead of Bayesian linear regression (as in ALPaCA), adaptation is done via Kalman filtering on the last layer of the network, allowing for task drift over time~\cite{west2006bayesian}.

Inputs at time $t$ (e.g., a history of data) are written $\x_t$ and outputs are written as $\y_t$. Our objective is to infer the probability of $\y_t$ conditional on $\x_t$. We write our predictive model as 
$\y_t = \Feat_{\w}(\x_t) \p_t + \ep_t$,
where $\Feat_{\w}(\cdot)$ is a matrix of neural network features parameterized by $\w$, $\p_t$ is a (time-varying) vector last layer, and $\ep_t$ is a zero mean Gaussian noise instantiation at time $t$ with covariance $\Sep$ (assumed independent across time). This differs from standard neural network regression in several ways. First, instead of a vector of features and a matrix last layer, we have a vector last layer and a matrix of features; this yields simpler inference within the model. Second, the last layer is assumed time-varying; we choose parameter dynamics
\begin{equation}
\p_{t+1} = A_{\w}(\x_{t}) \p_t + \bm{b}_{\w}(\x_t) + \pep_t
\end{equation}
to enable tractability of inference. We typically assume simple $A, \bm{b}$ in practice, such as the identity matrix for $A$. The term $\pep_t$ is a noise term, with variance $\Spep$. There are several alternate choices that can be made in this framework while retaining inferential tractability, discussed in depth in~\cite{Harrison2021}.

Inference within this model is as follows. Similar to Kalman filtering for state estimation, filtering consists of a prediction step---where the dynamics are applied to predict how the parameters change forward in time---and correction, where a measurement is used to update the estimate. 

The prediction step is 
\begin{equation}\label{eqn:pred_step}
\begin{aligned}
    \bar{\p}_{t+1|t} &= A_t \bar{\p}_{t|t} + \bm{b}_t\\
    S_{t+1|t} &= A_t S_{t|t} A_t + \Spep,
\end{aligned}
\end{equation}
where the subscript $t+1|t$ denotes the prediction for the value of the parameter at time $t+1$ made at time $t$. In the above, $A_t$ is shorthand for $A(\x_t)$, which we will use for other quantities. The terms $\bar{\p}$ and $S$ correspond to the mean and variance of the Gaussian prediction. 

The correction step is 
\begin{equation}\label{eqn:corr_step}
\begin{aligned}
    P_{t+1} &= \Feat_{t+1} S_{t+1|t} \Feat_{t+1}^\top + \Sep\\
    K_{t+1} &= S_{t+1|t} \Feat_{t+1}^\top P_{t+1}^{-1}\\
    \bm{e}_{t+1} &= \y_{t+1} - \Feat_{t+1} \bar{\p}_{t+1|t}\\
    \bar{\p}_{t+1|t+1} &= \bar{\p}_{t+1|t} + K_{t+1} \bm{e}_{t+1} \\
    S_{t+1|t+1} &= S_{t+1|t} - K_{t+1} \Feat_{t+1} S_{t+1|t}.
\end{aligned}
\end{equation}
From this framework, the predictive loss is the log likelihood of the prediction using mean and variance $\bar{\p}_{t+1|t}, S_{t+1|t}$. Critical to the meta-learning formulation, the parameters of the feature matrix, (possibly) the parameter dynamics matrices, and the noise covariances are trained by backpropagating through the iterated inference and prediction, using the log-likelihood over a segment of an episode as a loss. The model thus learns features that are capable of adapting to novel environments. %

\section{Problem Formulation}\label{sec:problem}

Our core problem formulation follows that of trajectory forecasting. Namely, we wish to generate plausible trajectory distributions for a time-varying number $N(t)$ of agents $A_1, \dots, A_{N(t)}$. At time $t$, given the state $\s^{(t)}_i \in \R^D$ of each agent (e.g., relative position, velocity, acceleration, and heading), their histories for the previous $H$ timesteps, and optional scene context $I^{(t)} \in \mathcal{I}$ (e.g., a local semantic map patch), we seek a distribution over each agents' future states for the next $T$ timesteps $\y_t = \s^{(t+1:t+T)}_{1,\dots,N(t)} \in \R^{T \times N(t) \times D}$, which we denote as $p(\y_t \mid \x_t)$, where $\x_t = (\s_{1,\dots,N(t)}^{(t-H:t)}, I^{(t)}) \in \R^{(H+1) \times N(t) \times D} \times \mathcal{I}$ contains both the agent state histories and (optional) scene context $I^{(t)}$.

At training time, we assume access to a dataset $\D_\mathcal{S} = \{\x_j, \y_j\}_{j=1}^{|\D_S|}$, collected from a set of source environments $\mathcal{S}$, and aim to obtain a model $p(\y \mid \x)$ that can be effectively deployed to a target environment $\mathcal{T}$. Since this work tackles methods for adaptation, we focus on the following two problem settings that reflect the types of domain transfer commonly considered in practice: \emph{Online}, where methods must adapt to \emph{streaming} data (i.e., agents' current states $\s_{1, \dots, N(t)}^{(t)}$ and optional scene context $I^{(t)}$ are observed at every timestep $t$), and
\emph{Offline}, where prediction methods have access to a small amount of \emph{already-collected} data $\D \subset \D_\mathcal{T}$ from the target environment $\mathcal{T}$ with which they can finetune before deployment.

\section{Generalizing Prediction Models Through Adaptive Meta-Learning}\label{sec:method}

\subsection{Architecture}

Our architecture takes an encoder-decoder structure, similar to that of Trajectron++~\cite{SalzmannIvanovicEtAl2020}. Inputs (e.g., agent histories $\x$, encoded scene context $I$) are mapped to an overall scene encoding vector $\mathbf{v}$
via an attentional graph-structured recurrent network.
The encoding $\mathbf{v}$ is then fed to the recurrent decoder, which maps $\mathbf{v}$ and the current hidden state $h_t$, as well as input states of the ego agent and other agents in the environment $\smash{\s^{(t)}_{1,\dots,N(t)}}$, to a distribution over actions taken by the ego agent. This action is then sampled and passed through the chosen dynamics model to generate the next state, which can then be fed back into the decoder at the next timestep.

Whereas Trajectron++ modeled multimodality via discrete latent variables~\cite{SalzmannIvanovicEtAl2020}, our model accounts for multimodality via controllable \textit{aleatoric} uncertainty, sampling, and the recurrent network dynamics. We parameterize both output features $\Feat_t$ and the predictive noise covariance $\Sigma_t$ as a function of the RNN hidden state. In particular, $\Feat_t = \Feat(h_t)$ and $\Sigma_t = \Sep(h_t)$, where $h_t$ is the decoder's hidden state. This parameterization of the noise covariance as a function of history is important: it captures irreducible aleatoric uncertainty as a function of state (or state history).

In contrast to epistemic uncertainty, aleatoric uncertainty is \textit{irreducible}. For example, there is always some uncertainty as to the intentions of a car at an intersection, which can not be reduced through further observation (in contrast to epistemic uncertainty).
In our model, this uncertainty will be unimodal for the first prediction step, but feeding output samples back through the recurrent decoder generates multimodality, yielding an overall multimodal output.
Crucially, this controllable aleatoric uncertainty pairs naturally with our adaptive last layer: the higher the irreducible uncertainty in a state transition, the less information can be extracted from it for reducing epistemic uncertainty.

\subsection{Multi-step Prediction and Loss}\label{sec:multistep_pred}
In this subsection, we detail our model's prediction procedure and loss computation. 

\textbf{Prediction.} For multi-step prediction, we begin by sampling from the last layer, via $\p^i_{t+1|t} \sim \N(\bar{\p}_{t+1|t}, S_{t+1|t})$ for samples $i = 1, \ldots, N$. We then set $\tau = 0$ and repeat:
\begin{itemize}
    \item Sample $\ac^i_{t+\tau| t} \sim \N(\Feat_{t+\tau}^i \bar{\p}^i_{t+\tau|t}, \Sigma^i_{t+\tau})$
    \item Compute next state $\s^i_{{t+\tau+1| t}} = f(\s^i_{{t+\tau| t}}, \ac^i_{t+\tau| t})$
    \item Set $h_{t+\tau+1}^i \gets \text{DecoderRNNCell}(\s^i_{{t+\tau+1 | t}}, h_{t+\tau})$
    \item Sample $\bar{\p}^i_{t+\tau+1|t} \sim \N(\bar{\p}^i_{t+\tau|t}, \Spep)$
    \item Set $\tau \gets \tau + 1$
\end{itemize}
Looping this procedure for $T$ steps will give $N$ samples of $T$-length forecasts. Gradients may be computed through the sampling via the reparameterization trick~\cite{KingmaWelling2013}.

\setlength{\tabcolsep}{3pt}
\begin{table*}[t]
\fontsize{7}{7}\selectfont
\centering
\caption{Average Displacement Error (m) obtained when training and evaluating methods across scenes in the ETH/UCY pedestrian datasets. A, B, C, D, and E denote ETH, Hotel, Univ, Zara1, and Zara2. Results on FDE can be found in~\cref{sec:supp_peds}.}

\vspace{-0.25cm}

\begin{tabular}{l|cccccccccccccccccccc|c}
\toprule
& A2B & A2C & A2D & A2E & B2A & B2C & B2D & B2E & C2A & C2B & C2D & C2E & D2A & D2B & D2C & D2E & E2A & E2B & E2C & E2D & AVG \\ \midrule
S-STGCNN~\cite{MohamedQianEtAl2020} & 1.83 & 1.58 & 1.30 & 1.31 & 3.02 & 1.38 & 2.63 & 1.58 & 1.16 & 0.70 & 0.82 & 0.54 & 1.04 & 1.05 & 0.73 & 0.47 & 0.98 & 1.09 & 0.74 & 0.50 & 1.22 \\
PECNet~\cite{MangalamGiraseEtAl2020} & 1.97 & 1.68 & 1.24 & 1.35 & 3.11 & 1.35 & 2.69 & 1.62 & 1.39 & 0.82 & 0.93 & 0.57 & 1.10 & 1.17 & 0.92 & 0.52 & 1.01 & 1.25 & 0.83 & 0.61 & 1.31\\
RSBG~\cite{SunJiangEtAl2020} & 2.21 & 1.59 & 1.48 & 1.42 & 3.18 & 1.49 & 2.72 & 1.73 & 1.23 & 0.87 & 1.04 & 0.60 & 1.19 & 1.21 & 0.80 & 0.49 & 1.09 & 1.37 & 1.03 & 0.78 & 1.38\\
Tra2Tra~\cite{XuRenEtAl2021} & 1.72 & 1.58 & 1.27 & 1.37 & 3.32 & 1.36 & 2.67 & 1.58 & 1.16 & 0.70 & 0.85 & 0.60 & 1.09 & 1.07 & 0.81 & 0.52 & 1.03 & 1.10 & 0.75 & 0.52 & 1.25\\
SGCN~\cite{ShiWangEtAl2021} & 1.68 & 1.54 & 1.26 & 1.28 & 3.22 & 1.38 & 2.62 & 1.58 & 1.14 & 0.70 & 0.82 & 0.52 & 1.05 & 0.97 & 0.80 & 0.48 & 0.97 & 1.08 & 0.75 & 0.51 & 1.22\\
T-GNN~\cite{XuWangEtAl2022} & 1.13 & 1.25 & 0.94 & 1.03 & 2.54 & 1.08 & 2.25 & 1.41 & \bfseries 0.97 & 0.54 & 0.61 & \bfseries 0.23 & \bfseries 0.88 & 0.78 & 0.59 & \bfseries 0.32 & \bfseries 0.87 & 0.72 & 0.65 & \bfseries 0.34 & 0.96\\ \midrule
$K_0$ & 0.45 & 0.78 & 0.68 & 0.59 & \bfseries 0.87 & 0.65 & \bfseries 0.49 & \bfseries 0.36 & 1.08 & \bfseries 0.46 & \bfseries 0.44 & 0.36 & 1.13 & 0.56 & 0.59 & 0.53 & 1.13 & 0.39 & 0.59 & 0.48 & 0.63\\
Ours & \bfseries 0.36 & \bfseries 0.64 & \bfseries 0.53 & \bfseries 0.45 & 0.90 & \bfseries 0.62 & 0.50 & \bfseries 0.36 & 1.11 & \bfseries 0.46 & 0.48 & 0.42 & 1.11 & \bfseries 0.52 & \bfseries 0.54 & 0.47 & 1.05 & \bfseries 0.34 & \bfseries 0.56 & 0.44 & \bfseries 0.59\\ \bottomrule
\end{tabular}
\label{tab:peds_d2d}
\vspace{-0.6cm}
\end{table*}

\textbf{Loss computation.} We propose to approximate the predictive density via kernel density estimation \cite{chen2017tutorial}. In particular, we aim to maximize
\begin{equation}
    \frac{1}{T} \sum_{\tau = t}^T \log \frac{1}{N} \sum_{i=1}^N \N(\s_{t+\tau+1}; \s^i_{t+\tau+1|t}, V^i_{t+\tau+1}),
\end{equation}
where $\s_{t+\tau+1}$ corresponds to the true state at time $t$ and $V^i_{t+\tau+1}$ corresponds to the covariance matrix used at time ${t+\tau+1}$ for particle $i$.
Note that this loss can be implemented stably as the log-sum-exp of the log Gaussian density, and that
the choice of $V^i_{t+\tau+1}$ is important. 
If $V^i_{t+\tau+1}$ is fixed across time, we risk later samples inducing extremely large variance in our gradient estimation. This is a natural consequence of the outcome space for the integrator dynamics being wider for long horizons than short horizons,
and the uncertainty metric in KDE should reflect this. Accordingly, we propose to integrate the uncertainty via
$V^i_{t+\tau+1} = \sum_{k = t}^{t+\tau+1} \Sigma^i_{k}$.
In the case where the dynamics are a single integrator (a common choice in Trajectron++~\cite{SalzmannIvanovicEtAl2020}), this captures the state uncertainty induced by aleatoric control uncertainty exactly, and thus it naturally scales in time.

\begin{figure}[t]
  \centering
  \includegraphics[width=1.0\linewidth]{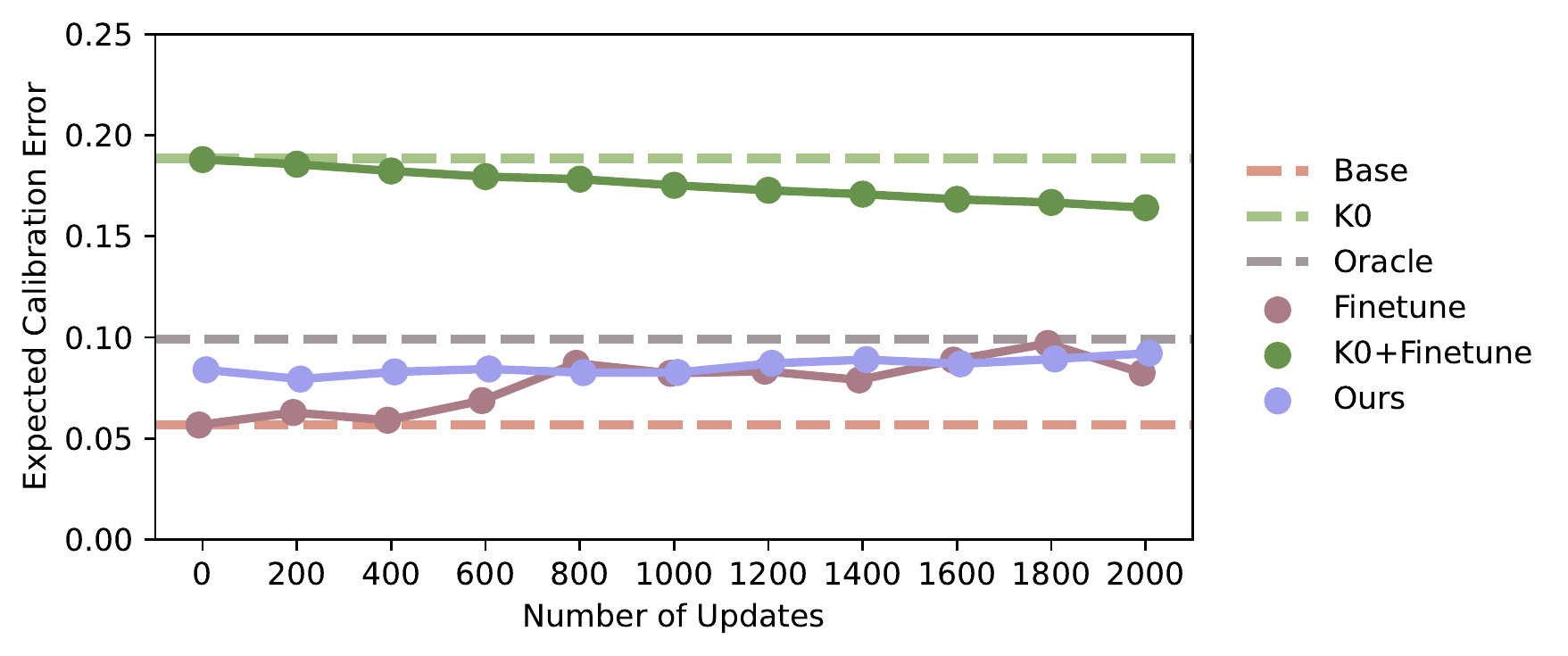}
  
  \vspace{-0.4cm}
  
  \caption{[Zara1 $\rightarrow$ Hotel] Our method learns a well-calibrated prior, and maintains its calibration as it observes more data. Although gradient-based finetuning is initially well-calibrated, it overfits to the observed data and produces a less-calibrated model. Lower is better.}
  \label{fig:peds_offline_ece}
  \vspace{-0.5cm}
\end{figure}

\subsection{Adaptation}\label{sec:adaptation}

While we have described a multi-step prediction formulation, we will only use one step predictions for adaptation. Because the first prediction is unimodal and Gaussian, adaptation may be performed using exact inference on the last layer. While this choice reduces the maximum theoretical performance of our approach, 
instead of \emph{approximately} adapting our model to the multi-step prediction error, we perform \emph{exact} inference based on the single step predictions. Most importantly, this trade-off enables high execution frequencies due to the simplicity of the update and high data efficiency of exact inference. Further discussion of computational complexity is available in~\cref{sec:supp_computation}. 

{\bf Online Adaptation.} In the online adaptation setting, an agent makes a probabilistic prediction, for which the initial prediction for all particles is the same. The mean and variance of this prediction are used to update the last layer, which corresponds exactly to the update equations described in Section \ref{sec:bg_adaptive}. Critically, adaptation in this setting is done \textit{per agent} (as each agent has only local information) and with temporally correlated data. 

{\bf Offline Adaptation.} In offline adaptation, a small dataset from the target environment is assumed to be provided. This dataset, which consists of several different agents' state information, is then used to adapt the last layer via the same adaptation mechanism. 

{\bf Combining Last Layer Adaptation with Gradient-Based Finetuning.}
We can also combine our approach with gradient-based finetuning for even more efficient and effective domain transfer in the offline setting. In particular, we first perform $M$ steps of last layer adaptation and then switch to gradient-based finetuning for future updates. This sequential update scheme leverages the fast initial adaptation of our last layer exact inference, while exploiting the high capacity and strong performance of gradient-based fine-tuning \cite{howard2018universal}. This combination of exact inference for efficiency combined with gradient-based fine-tuning for capacity has been exploited previously~\cite{WillesHarrisonEtAl2021}, although not purely sequentially. 

\section{Experiments}\label{sec:expts}

We evaluate our method\footnote{Code and models are available online at \url{https://github.com/NVlabs/adaptive-prediction}.} on a variety of transfer scenarios from real-world pedestrian and vehicle data, comparing against prior works and ablations of our method on multiple deterministic and probabilistic metrics.
Our approach was implemented in PyTorch~\cite{PaszkeGrossEtAl2017} and all datasets were interfaced through \texttt{trajdata}~\cite{Ivanovic2022}, a recently-released unified interface to trajectory forecasting datasets.

{\bf Datasets.} We use the ETH~\cite{PellegriniEssEtAl2009} and UCY~\cite{LernerChrysanthouEtAl2007} pedestrian datasets, as well as the nuScenes~\cite{CaesarBankitiEtAl2019} and Lyft Level 5~\cite{HoustonZuidhofEtAl2020} autonomous driving datasets. The ETH and UCY datasets are a standard benchmark in the field, comprised of pedestrian trajectories captured at 2.5 Hz ($\Delta t = 0.4s$) in Zurich and Cyprus, respectively. In total, there are 5 sets of data from 4 unique environments, containing a total of 1536 pedestrians. %

nuScenes is a large-scale autonomous driving dataset comprised of 1000 scenes collected in Boston and Singapore. Each scene is annotated at 2 Hz ($\Delta t = 0.5s$) and is $20s$ long. Lyft Level 5 is comprised of 170K scenes collected in Palo Alto, each of which are annotated at 10 Hz ($\Delta t = 0.1s$) and roughly $25s$ long.

{\bf Ablations and Oracle.} We compare against the following five ablations as well as an oracle:
(1)~\emph{Base} is the original Trajectron++~\cite{SalzmannIvanovicEtAl2020} model without any of our adaptive architecture;
(2)~\emph{Finetune} is Base combined with gradient-based finetuning for adaptation, using an initial learning rate $10\times$ smaller than Base, as is common in practice.
(3)~$K_0$ is our method without any adaptive training, i.e., always setting $\bar{\p}_{t|t} = \bar{\p}_{0|0}$ and $S_{t|t} = S_{0|0}$ in \cref{eqn:pred_step} and not applying corrections;
(4)~\emph{$K_0$+Finetune} is the $K_0$ baseline combined with gradient-based finetuning. Note, only the last layer is finetuned (the rest of the model is frozen). In essence, this ablation compares using gradient descent for adaptation instead of the correction step in \cref{eqn:corr_step}; and
(5)~\emph{Ours+Finetune} is our full method combined with gradient-based finetuning, as described in \cref{sec:adaptation}. In our experiments, we switch to gradient-based finetuning after $M = 100$ Bayesian last layer updates. 
Finally, since our experiments focus on transferring models from one environment to another, we also show the performance of an oracle, i.e., the base Trajectron++~\cite{SalzmannIvanovicEtAl2020} model trained on data from the target environment, representing ideal transfer performance. Additional baselines will be described in their respective sections.

{\bf Metrics.} We evaluate prediction accuracy with a variety of deterministic and probabilistic metrics: \emph{Average/Final Displacement Error (ADE/FDE)}: mean/final $\ell_2$ distance between the ground truth (GT) and predicted trajectories; \emph{minADE$_{5/10}$}: ADE between the GT and best of 5 or 10 samples, respectively; and \emph{Negative Log-Likelihood (NLL)}: mean NLL of the GT trajectory under the predicted distribution. In addition to prediction accuracy, we also evaluate methods' predictive calibration by computing their Expected Calibration Error (ECE) \cite{ovadia2019can}, which compares how closely a model's predictive uncertainty matches its empirical uncertainty, as defined by the fraction of GT future positions lying within specified probability thresholds. 
\subsection{Pedestrian Data}

\begin{figure}[t]
  \centering
  \includegraphics[width=1.0\linewidth]{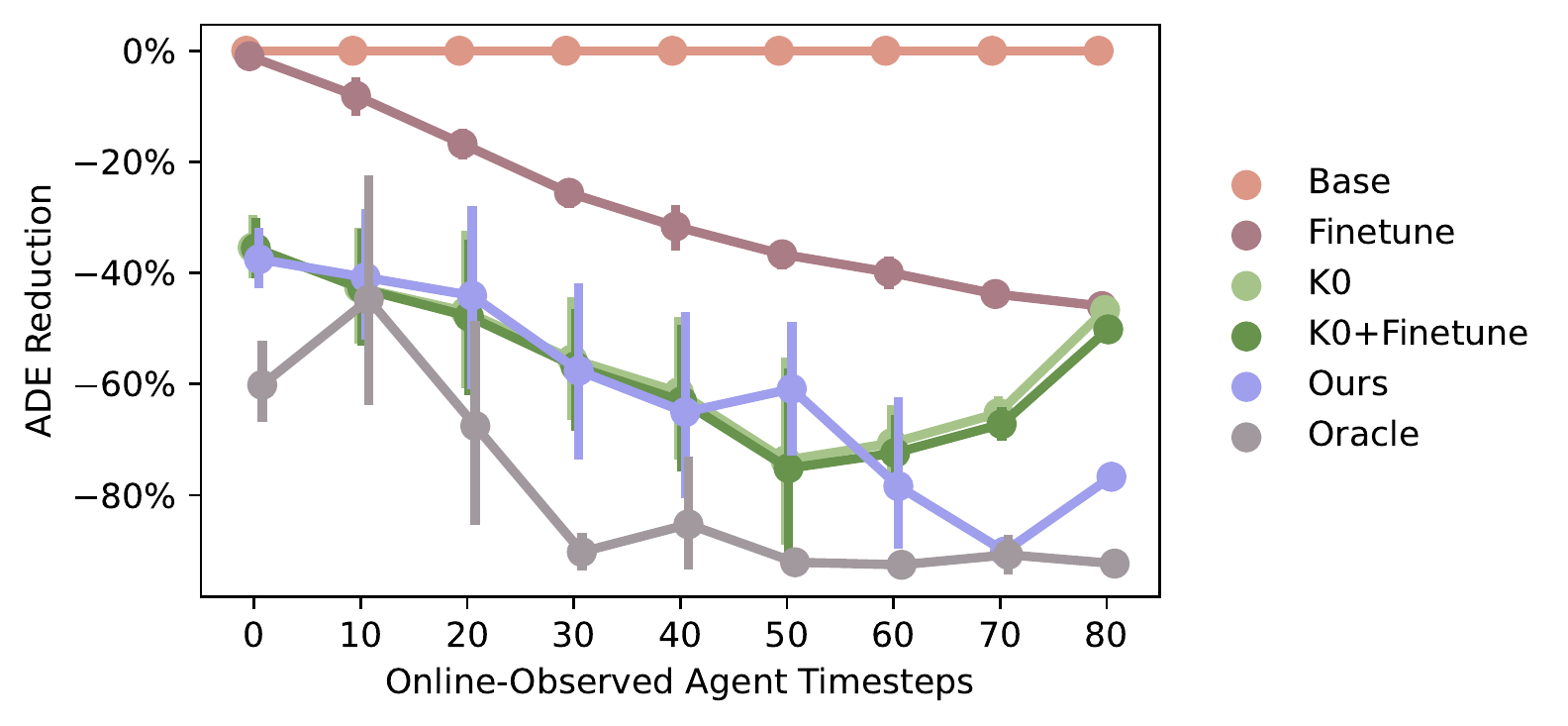}
  
  \vspace{-0.4cm}
  
  \caption{[Zara1 $\rightarrow$ Hotel] Our method's prediction accuracy improves rapidly as it observes data online, significantly outperforming na\"ive transfer and other ablations, even matching the oracle. Error bars are 95\% CIs, lower is better. Results on additional metrics can be found in~\cref{sec:supp_peds}.}
  \label{fig:peds_online}
  \vspace{-0.3cm}
\end{figure}

\begin{figure}[t]
  \centering
  \includegraphics[width=1.0\linewidth]{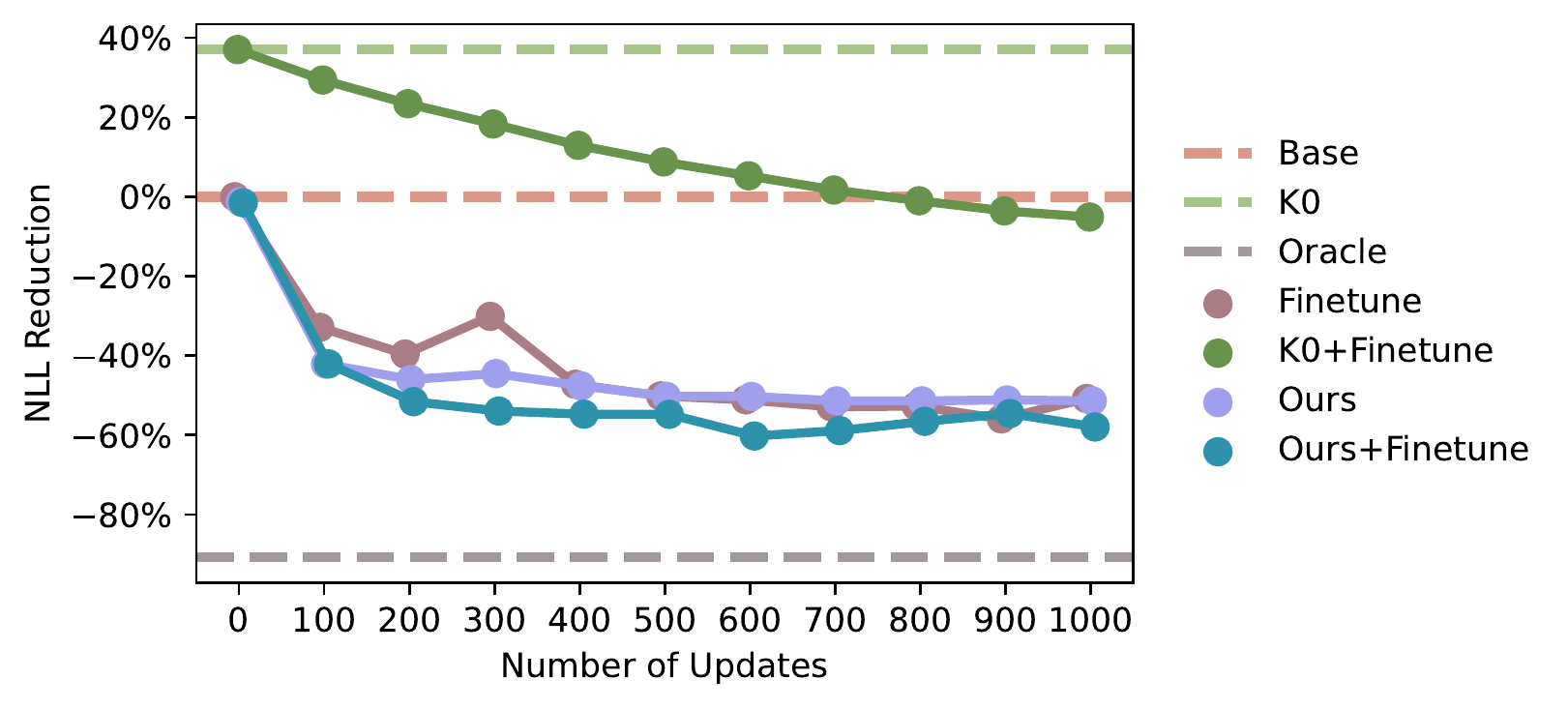}
  
  \vspace{-0.4cm}
  
  \caption{[nuScenes $\rightarrow$ Lyft] Even in the face of significant domain shift (i.e., $5\times$ timestep frequency and different map annotations), our method's predictions improve smoothly with more data, outperforming whole-model fine-tuning after only 200 data samples. Lower is better. Results on additional metrics can be found in~\cref{sec:supp_lyft}.}
  \label{fig:lyft_offline}
  \vspace{-0.4cm}
\end{figure}

\begin{figure}[t]
  \centering
  \includegraphics[width=1.0\linewidth]{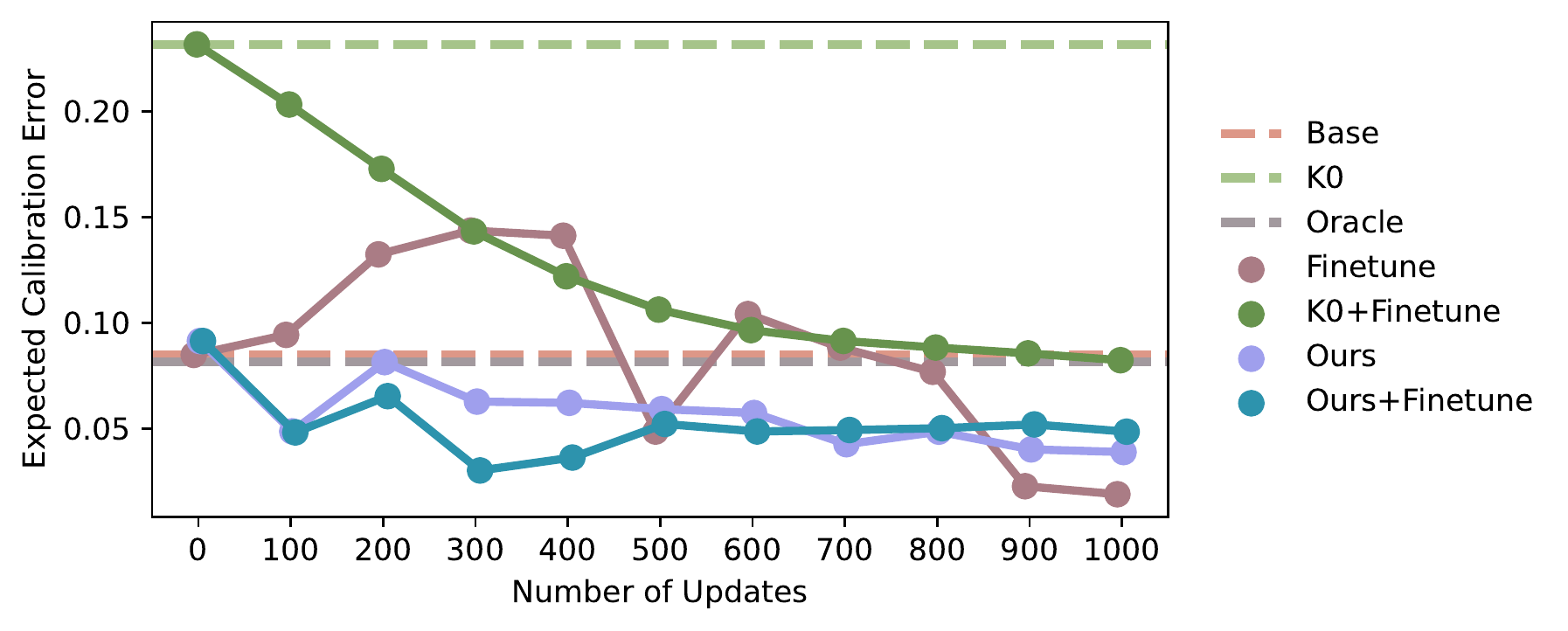}

  \vspace{-0.4cm}
  
  \caption{[nuScenes $\rightarrow$ Lyft] While our method's calibration already improves faster than baselines, combining it with gradient-based finetuning further accelerates improvement. Gradient-based finetuning alone yields unpredictable changes in calibration. Lower is better.}
  \label{fig:lyft_offline_ece}
  \vspace{-0.3cm}
\end{figure}

\begin{figure}[t]
  \centering
  \includegraphics[width=1.0\linewidth]{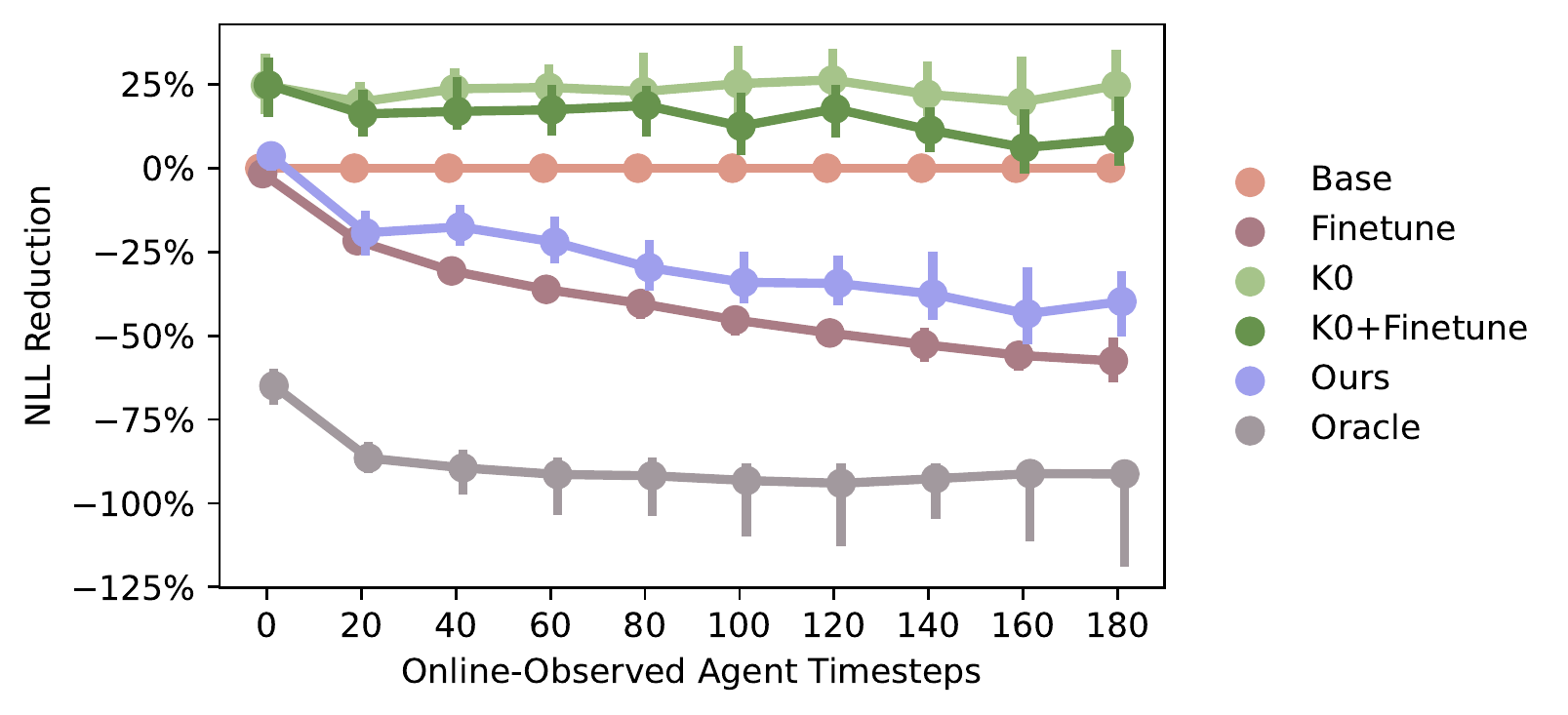}

  \vspace{-0.5cm}
  
  \caption{[nuScenes $\rightarrow$ Lyft] Our method's online NLL reduction tracks closely to that of whole-model finetuning, while being much less computationally expensive. Error bars are 95\% CIs, lower is better. Results on additional metrics can be found in~\cref{sec:supp_lyft}.}
  \label{fig:lyft_online}
  \vspace{-0.7cm}
\end{figure}

\begin{figure*}[t]
  \centering
  \includegraphics[align=b,height=0.16\linewidth]{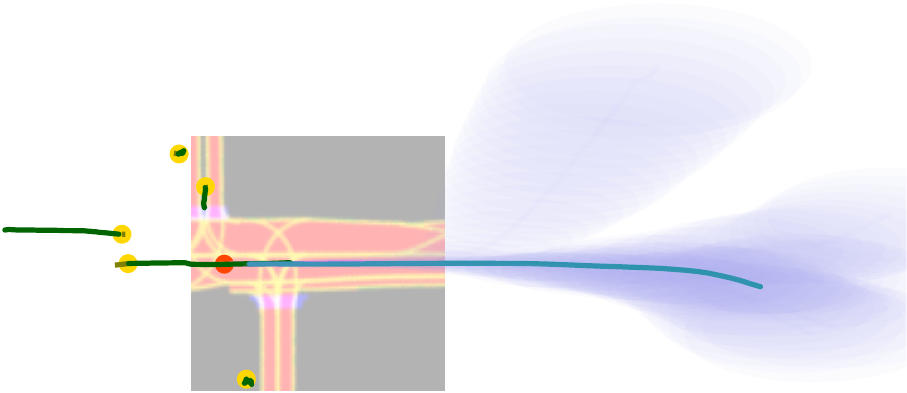}
  \includegraphics[align=b,height=0.1\linewidth]{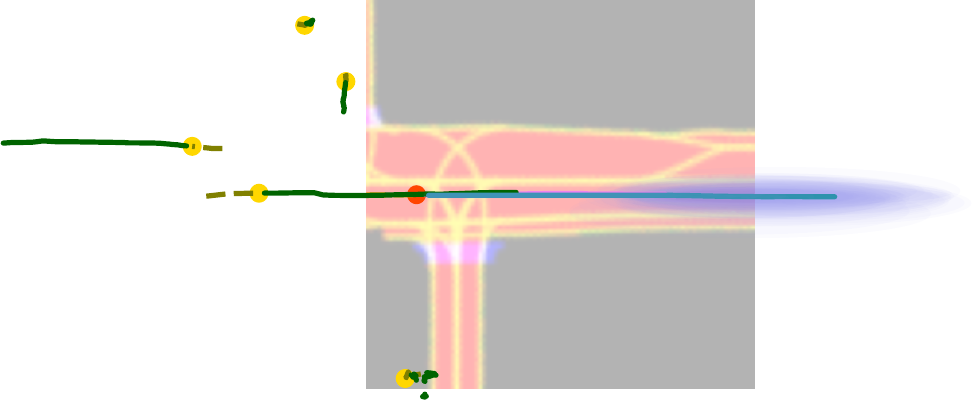}
  \includegraphics[align=b,height=0.1\linewidth]{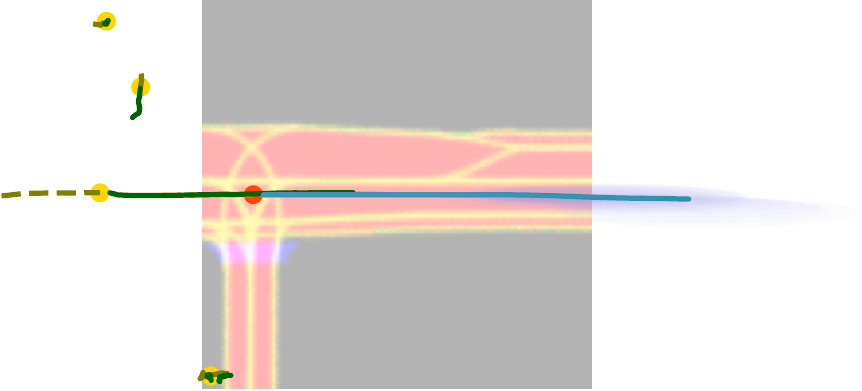}
  \includegraphics[align=b,height=0.1\linewidth]{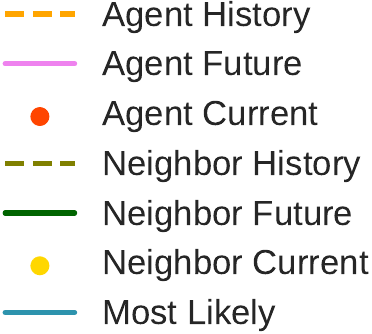}
  
  \vspace{-0.3cm}
  
  \caption{[nuScenes $\rightarrow$ Lyft] \textbf{Left}: Initially, before making any adaptive steps, our model's prior predictions are uncertain and spread out (primarily due to the significant domain shift between nuScenes and Lyft). After observing only $0.5s$ of data online (\textbf{Middle}), it has adapted to the environment and its predictions better match the GT future. Finally, after observing $1s$ of data (\textbf{Right}), our method has tightly clustered its spatial probability around the GT.}
  \label{fig:lyft_online_qual}
  \vspace{-0.5cm}
\end{figure*}

\begin{figure}[t]
  \centering
  \includegraphics[align=c,width=0.3\linewidth]{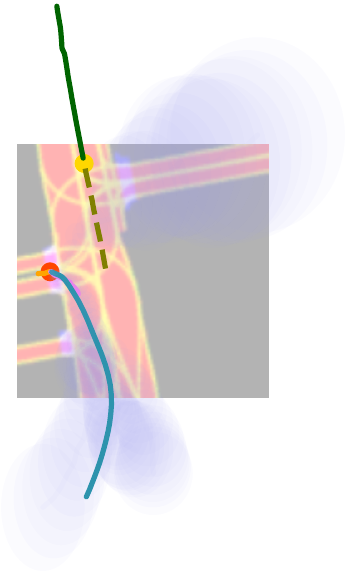}
  \includegraphics[align=c,height=0.22\linewidth]{figures/qual_legend.pdf}
  
  \caption{[nuScenes $\rightarrow$ Lyft] Our multi-step sampling strategy yields multimodal predictions. 5 samples are shown here, as well as the most-likely prediction.}
  \label{fig:lyft_multimodal_qual}
  \vspace{-0.7cm}
\end{figure}

{\bf Evaluation Protocol.} As in recent prior work~\cite{XuWangEtAl2022}, we treat each of the 5 scenes in the ETH and UCY datasets as individual source domains $\mathcal{S} \in \{$\texttt{ETH}, \texttt{Hotel}, \texttt{Univ}, \texttt{Zara1}, and \texttt{Zara2}$\}$, and use the other 4 scenes as our target domains $\mathcal{T} \neq \mathcal{S}$, yielding 20 cross-domain pairings. To ensure that we are purely evaluating adaptation, we restrict models' training sets to \emph{only} be their source domain's training split $\D_{\mathcal{S},\textrm{train}}$ (leaving the source domain's validation set $\D_{\mathcal{S},\textrm{val}}$ for hyperparameter tuning) and evaluate their performance on the entire target dataset $\D_{\mathcal{T}}$. This is a notable difference from the evaluation protocol proposed in~\cite{XuWangEtAl2022}, where prediction methods train on the source domain's training split $\D_{\mathcal{S},\textrm{train}}$ \emph{as well as the target domain's validation split} $\D_{\mathcal{T},\textrm{val}}$. This complicates measuring a model's capability for domain adaptation, as it has already seen data from the target environment during training. 
In the rest of this section, up to $H=8$ frames (current and history) are observed and the next $T=12$ frames are predicted, as in prior works~\cite{XuWangEtAl2022}.

{\bf Scene-to-Scene Transfer.} In this setting, we train on a source scene X and directly transfer to a target scene Y (denoting the pair as ``X2Y"), without any online or offline adaptation to fairly compare with prior work.
We additionally compare to a large variety of state-of-the-art approaches, namely S-STGCNN~\cite{MohamedQianEtAl2020}, PECNet~\cite{MangalamGiraseEtAl2020}, RSBG~\cite{SunJiangEtAl2020}, Tra2Tra~\cite{XuRenEtAl2021}, SGCN~\cite{ShiWangEtAl2021}, and T-GNN~\cite{XuWangEtAl2022}. 
As can be seen in \cref{tab:peds_d2d}, our $K_0$ ablation significantly outperforms all baseline approaches on the vast majority of transfer settings, showing that our prior and its analytical propagation of uncertainty already yields strong performance. Our full method further improves upon $K_0$ due to its additional conditioning on observed history (i.e., using \cref{eqn:corr_step} with the agent's $H$-length trajectory history in each data sample).

In the ETH/UCY datasets, scenes primarily differ due to their geographic regions (e.g., Zurich and Cyprus) and layout (i.e., a university campus and an urban sidewalk). Accordingly, transfer asymmetries can be seen in \cref{tab:peds_d2d}, e.g., training on the ETH - Univ scene and transferring to an urban sidewalk in Cyprus (A2D and A2E) yields much lower errors than the reverse direction (D2A and E2A).

{\bf Offline Calibration.} \cref{fig:peds_offline_ece} demonstrates that our method learns a well-calibrated prior (Ours at 0 updates), and maintains its calibration as it observes more data offline. Further, while gradient-based finetuning starts well-calibrated, it overfits to the observed data and yields an overall less-calibrated model. The oracle is poorly calibrated because Trajectron++~\cite{SalzmannIvanovicEtAl2020} tends to be overconfident when predicting pedestrians (detailed calibration results showing this can be found in~\cref{sec:supp_peds}). Accurate calibration is critical to designing closed-loop autonomy stacks that are not over- or under-conservative in their decision making, and is thus vital for safe autonomy. 

{\bf Online Adaptation.} Focusing on the transfer from UCY-Zara1 to ETH-Hotel (D2B), \cref{fig:peds_online} shows that our method's median ADE reduction rapidly approaches the oracle as more of an agent's trajectory is observed online. Both Finetune and $K_0$+Finetune are slow to adapt online, with Finetune only reducing ADE by 40\% in the same time our method achieves 90\%+ reductions and $K_0$+Finetune barely differing from the non-adaptive $K_0$. Additional results on the other metrics can be found in~\cref{sec:supp_peds}.

\subsection{Autonomous Driving Data}

{\bf Evaluation Protocol.} We treat nuScenes as the source domain and Lyft Level 5 as the target domain. In particular, we train models on the nuScenes prediction challenge \texttt{train} split, tune hyperparameters on the \texttt{train\_val} split, and evaluate methods' capability to adapt to the entire Lyft Level 5 \texttt{sample} split. This is a \emph{much} more challenging domain transfer problem than in the ETH/UCY datasets, since the nuScenes and Lyft datasets feature different underlying map annotations (e.g., nuScenes does not annotate lanes through intersections whereas Lyft does) and data frequencies (2 Hz vs 10 Hz), in addition to being collected in diverse cities with unique road geometries. In the rest of this section, up to $2s$
of history are observed and the next $6s$
are predicted, as in the nuScenes prediction challenge.

{\bf Offline Transfer.} As can be seen in \cref{fig:lyft_offline}, our method rapidly reduces NLL, outperforming Finetune after receiving only 200 updates.
Ours+Finetune further improves upon our method, showing that gradient-based finetuning is a complementary and performant addition to our last layer adaptation scheme for offline transfer.

{\bf Offline Calibration.} \cref{fig:lyft_offline_ece} confirms that our approach's predictive calibration improves with more data. Switching to gradient-based finetuning after 100 last-layer adaptation steps significantly accelerates improvement. Gradient-based finetuning alone yields unpredictable changes in calibration.

{\bf Online Adaptation.} \cref{fig:lyft_online} reinforces that our method's performance rapidly improves as it sequentially observes data online. While whole-model finetuning (Finetune) is slightly more performant, it is not real-time feasible, highlighting our method's performance as it tracks closely to Finetune while being significantly less computationally expensive.

{\bf Qualitative Results.} \cref{fig:lyft_online_qual} shows our model's adaptation visually. Before seeing any data, our model's predictions are spread out, uncertain, and overshoot the GT (due to the nuScenes-Lyft data frequency mismatch). After observing only $0.5s$ of data, our model has adapted to the data frequency mismatch and its predictions better match the GT future. Finally, after observing $1s$ of data, our method has tightly clustered its spatial probability around the GT, yielding an accurate prediction with confident associated uncertainty. \cref{fig:lyft_multimodal_qual} shows that our proposed sampling scheme yields diverse, multimodal predictions in an unusual intersection in Palo Alto.

\section{Discussion and Conclusion}

In this work, we have presented a model that combines the strength of recurrent models with adaptive, optimization-based meta-learning. This combination both adapts efficiently and yields strong performance. Our approach has shown particularly strong results for the one-to-one transfer setting. One strength of the approach presented in this paper---which we do not address due to space constraints---is the ability to do effective many-to-one transfer, in which many training datasets are available. This setting is reflective of deploying in a new geographical location, where the full training dataset of many other cities is available for pre-training. We anticipate that further training data diversity will result in monotonic improvement to performance, yielding both better calibrated priors and more expressive learned features. Indeed, careful evaluation of the scaling performance with respect to meta-training data is an important direction of future work.

\section*{Acknowledgments}
We thank Nikolai Smolyanskiy and the rest of the NVIDIA AV Prediction team, John Willes, and Brian Ichter for helpful feedback on this paper. We also thank Anna M\'{e}sz\'{a}ros and Julian F.~Schumann for improving our ECE implementation.

\bibliographystyle{IEEEtran}
\bibliography{ASL_papers,main}

\clearpage
\newpage

\appendix

\setlength{\tabcolsep}{3pt}
\begin{table*}[t]
\fontsize{7}{7}\selectfont
\centering
\caption{Final Displacement Error (m) obtained when training and evaluating methods across scenes in the ETH/UCY pedestrian datasets. A, B, C, D, and E denote ETH, Hotel, Univ, Zara1, and Zara2.}

\vspace{-0.25cm}

\begin{tabular}{l|cccccccccccccccccccc|c}
\toprule
& A2B & A2C & A2D & A2E & B2A & B2C & B2D & B2E & C2A & C2B & C2D & C2E & D2A & D2B & D2C & D2E & E2A & E2B & E2C & E2D & AVG \\ \midrule
S-STGCNN~\cite{MohamedQianEtAl2020} & 3.24 & 2.86 & 2.53 & 2.43 & 5.16 & 2.51 & 4.86 & 2.88 & 2.30 & 1.34 & 1.74 & 1.10 & 2.21 & 1.99 & 1.41 & 0.88 & 2.10 & 2.05 & 1.47 & 1.01 & 2.30 \\
PECNet~\cite{MangalamGiraseEtAl2020} & 3.33 & 2.83 & 2.53 & 2.45 & 5.23 & 2.48 & 4.90 & 2.86 & 2.22 & 1.32 & 1.68 & 1.12 & 2.20 & 2.05 & 1.52 & 0.88 & 2.10 & 1.84 & 1.45 & 0.98 & 2.29 \\
RSBG~\cite{SunJiangEtAl2020} & 3.42 & 2.96 & 2.75 & 2.50 & 5.28 & 2.59 & 5.19 & 3.10 & 2.36 & 1.55 & 1.99 & 1.37 & 2.28 & 2.22 & 1.77 & 0.97 & 2.19 & 2.29 & 1.81 & 1.34 & 2.50 \\
Tra2Tra~\cite{XuRenEtAl2021} & 3.29 & 2.88 & 2.66 & 2.45 & 5.22 & 2.50 & 4.89 & 2.90 & 2.29 & 1.33 & 1.78 & 1.09 & 2.26 & 2.12 & 1.63 & 0.92 & 2.18 & 2.06 & 1.52 & 1.17 & 2.34 \\
SGCN~\cite{ShiWangEtAl2021} & 3.22 & 2.81 & 2.52 & 2.40 & 5.18 & 2.47 & 4.83 & 2.85 & 2.24 & 1.32 & 1.71 & 1.03 & 2.23 & 1.90 & 1.48 & 0.97 & 2.10 & 1.95 & 1.52 & 0.99 & 2.29 \\
T-GNN~\cite{XuWangEtAl2022} & 2.18 & 2.25 & 1.78 & 1.84 & 4.15 & 1.82 & 4.04 & 2.53 & \bfseries 1.91 & 1.12 & 1.30 & 0.87 & \bfseries 1.92 & 1.46 & 1.25 & \bfseries 0.65 & \bfseries 1.86 & 1.45 & 1.28 & \bfseries 0.72 & 1.82 \\ \midrule
$K_0$ & 0.94 & 1.72 & 1.51 & 1.29 & \bfseries 1.81 & 1.33 & \bfseries 1.02 & \bfseries 0.75 & 2.17 & \bfseries 0.91 & \bfseries 0.94 & \bfseries 0.81 & 2.18 & \bfseries 0.98 & 1.20 & 1.03 & 2.12 & 0.68 & 1.22 & 1.03 & 1.28 \\
Ours & \bfseries 0.69 & \bfseries 1.34 & \bfseries 1.12 & \bfseries 0.91 & 1.85 & \bfseries 1.27 & 1.05 & 0.76 & 2.26 & 0.92 & 1.03 & 0.94 & 2.23 & 1.02 & \bfseries 1.17 & 1.00 & 2.07 & \bfseries 0.63 & \bfseries 1.19 & 0.94 & \bfseries 1.22 \\ \bottomrule
\end{tabular}
\label{tab:supp_peds_d2d}
\vspace{-0.6cm}
\end{table*}

\begin{figure}[t]
  \centering
  \includegraphics[width=1.0\linewidth]{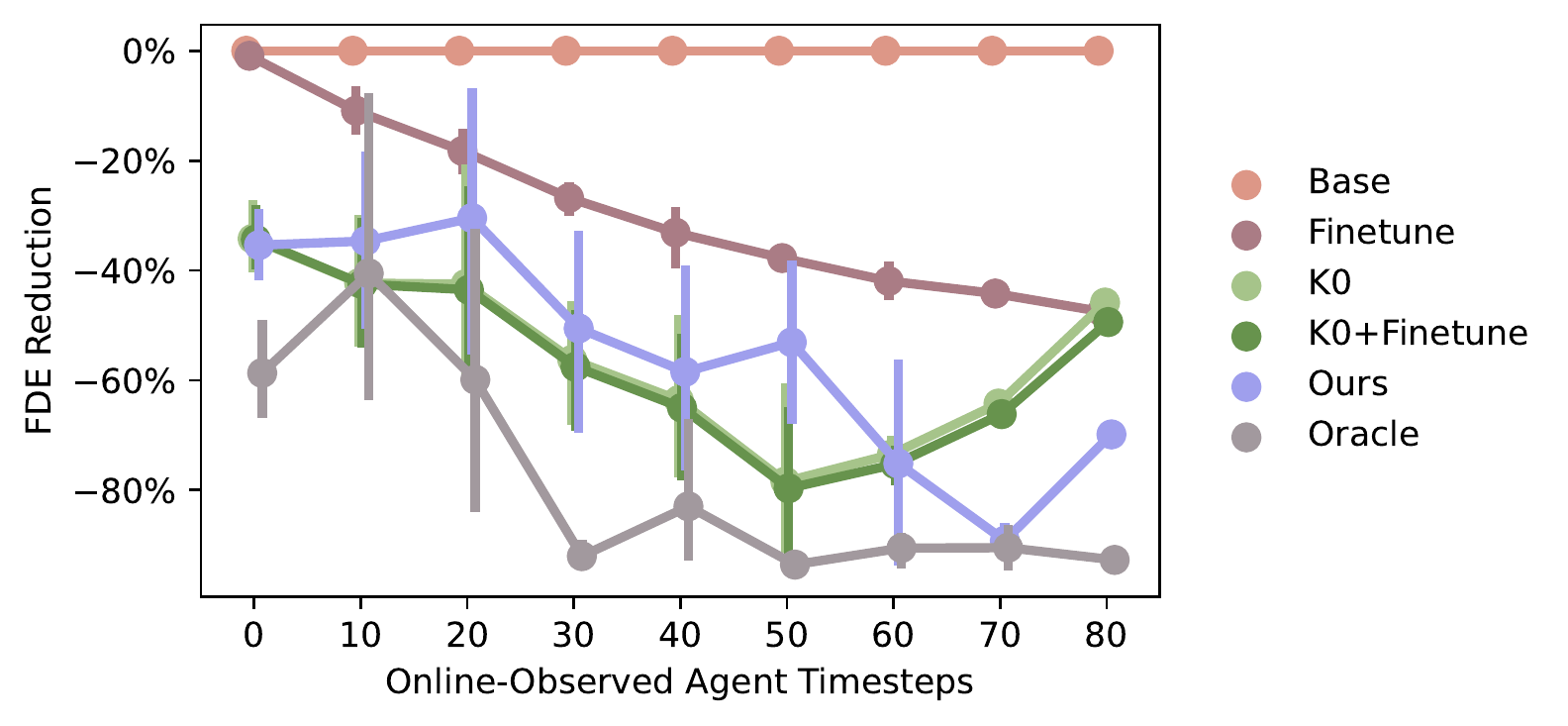}
  \includegraphics[width=1.0\linewidth]{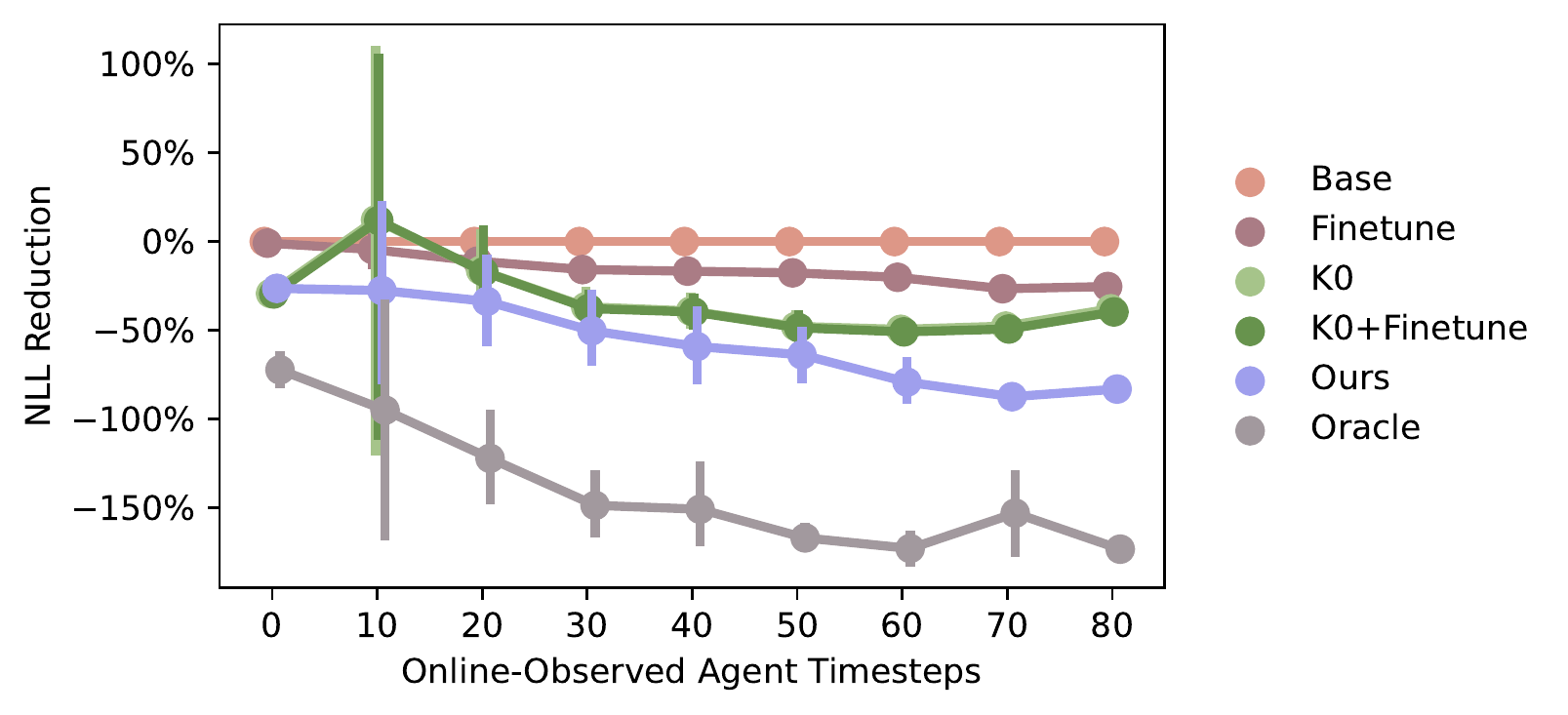}
  \caption{[Zara1 $\rightarrow$ Hotel] Our method's mean FDE and NLL reductions replicate the ADE results from \cref{fig:peds_online}, significantly outperforming na\"ive transfer and other ablations, even matching the oracle on FDE. Error bars are 95\% CIs, lower is better.}
  \label{fig:supp_peds_online}
\end{figure}

\begin{figure*}[t]
  \centering
  \includegraphics[height=0.18\linewidth]{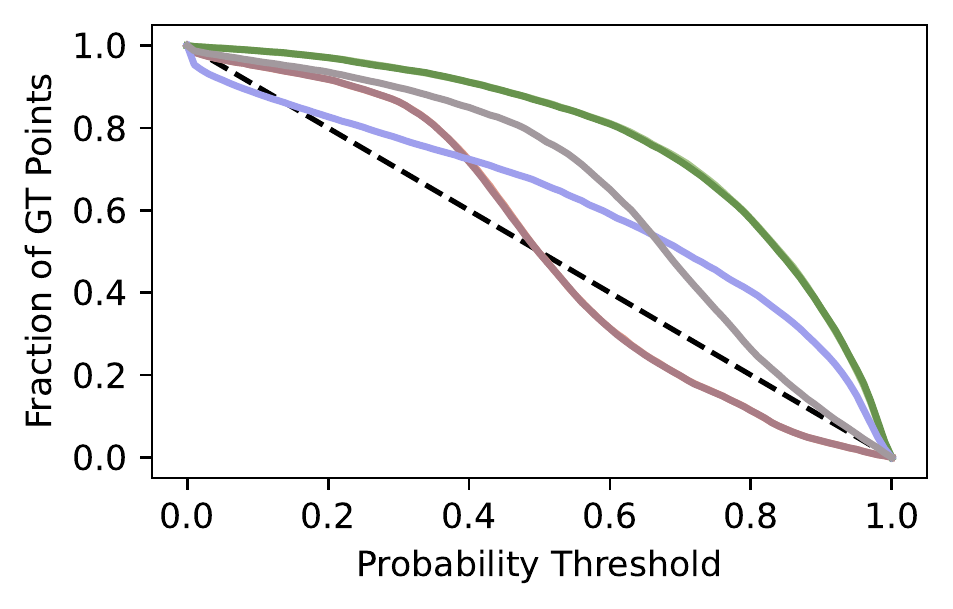}
  \includegraphics[height=0.18\linewidth]{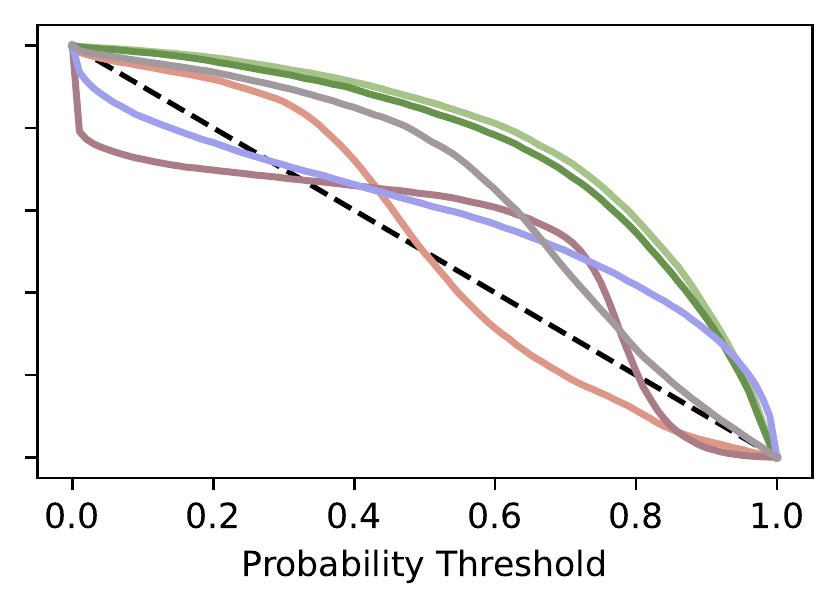}
  \includegraphics[height=0.18\linewidth]{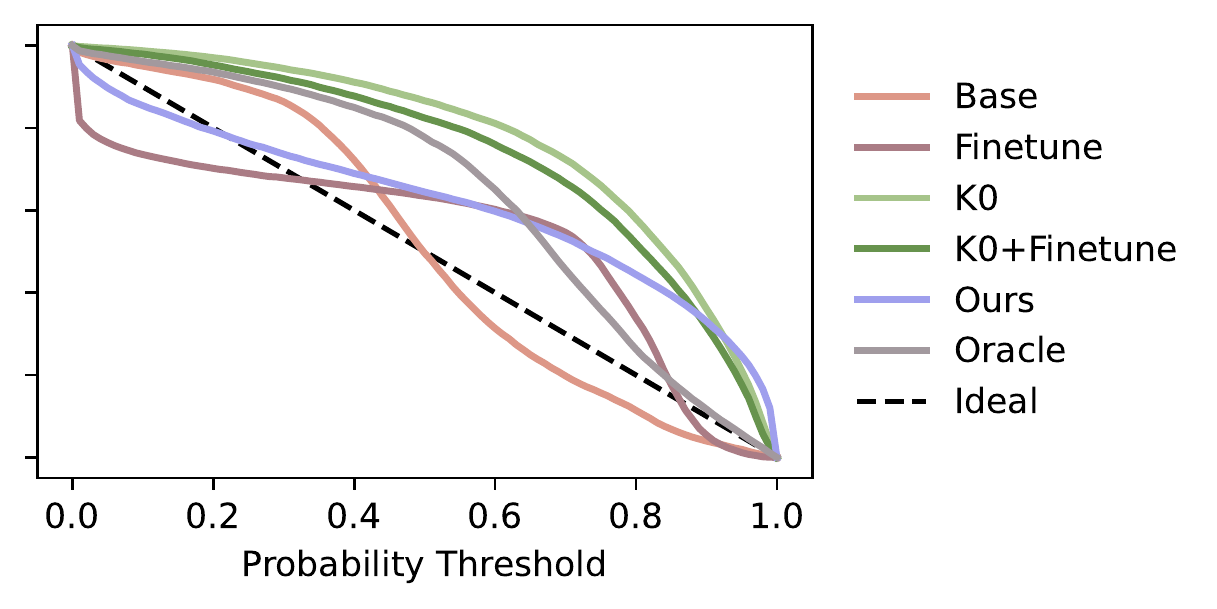}
  \caption{[Zara1 $\rightarrow$ Hotel] Detailed calibration plots showing the fraction of GT future positions lying within specified probability thresholds after observing 0, 1000, and 2000 data samples in the offline setting. Our method is closest to the ideal line, and stays close as more data points are observed. The area below the ideal line signifies underconfidence. Accordingly, the area above the ideal line signifies overconfidence (especially visible for the oracle). Computing the area between each line to the ideal line yields the ECE values after 0, 1000, and 2000 updates in \cref{fig:peds_offline_ece}.}
  \label{fig:supp_peds_offline_calib}
\end{figure*}

\begin{figure}[t]
  \centering
  \includegraphics[width=1.0\linewidth]{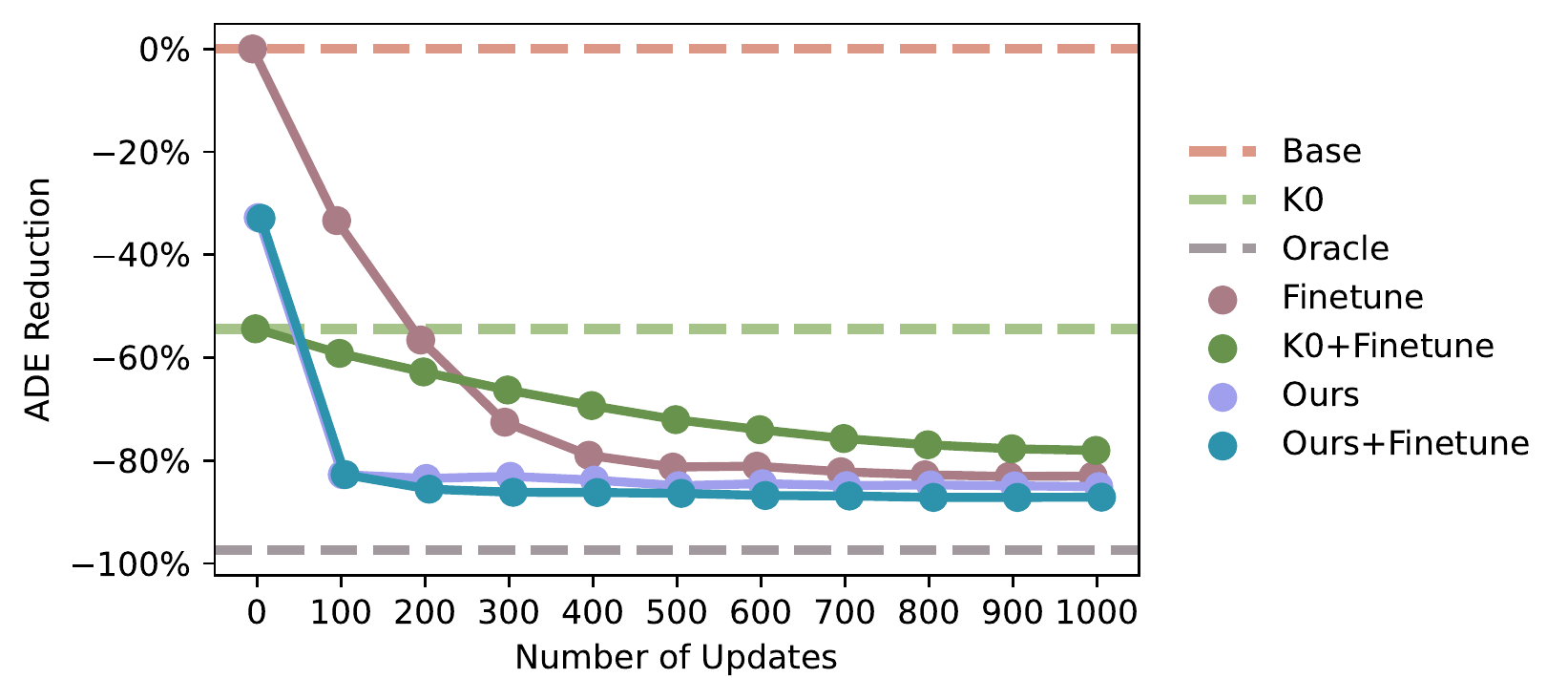}
  \includegraphics[width=1.0\linewidth]{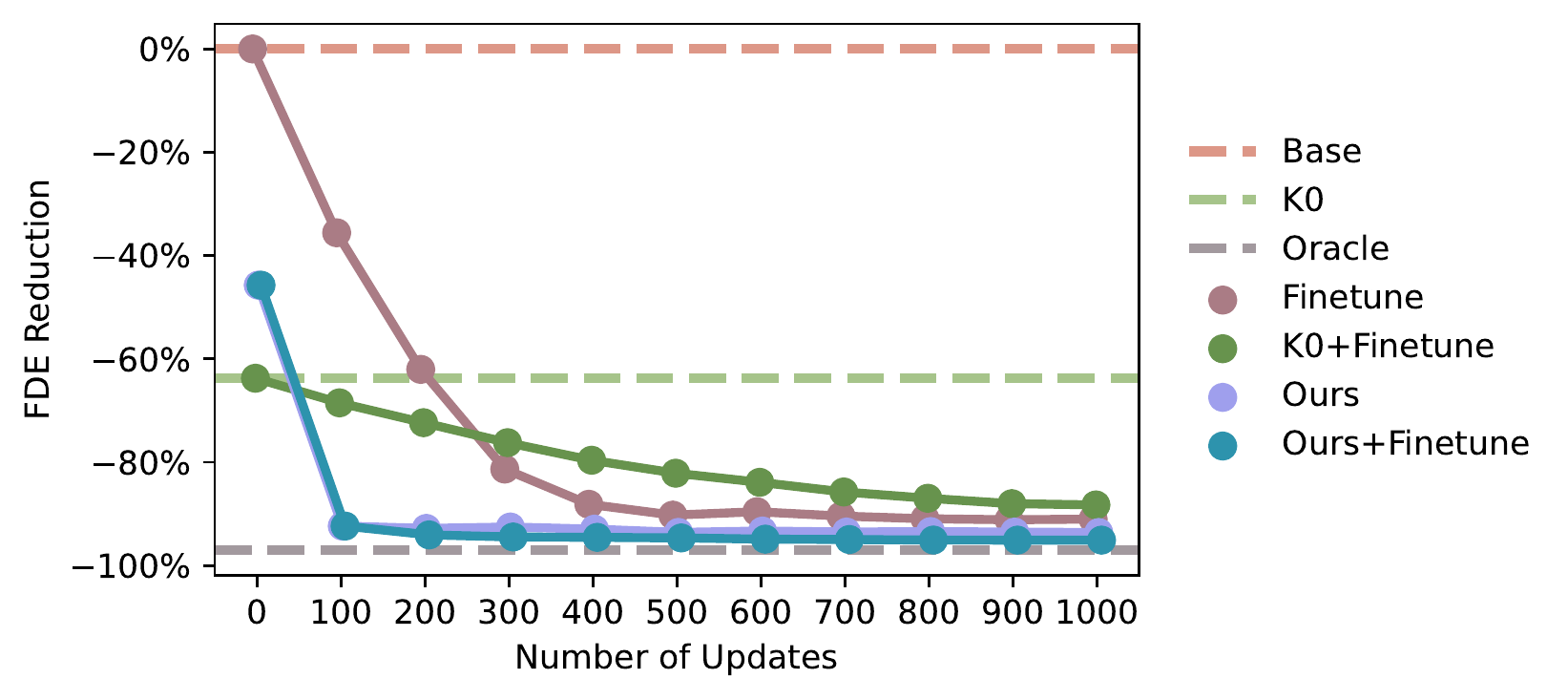}
  \includegraphics[width=1.0\linewidth]{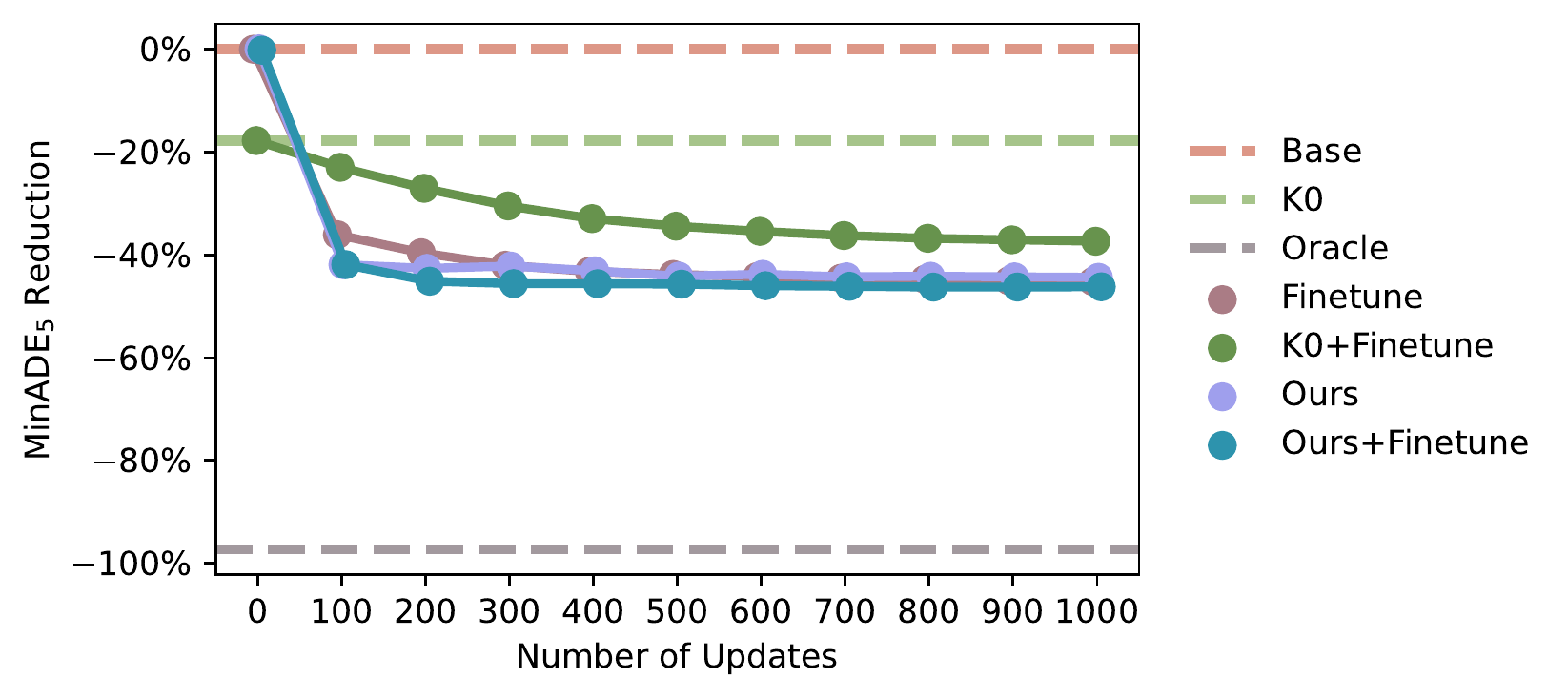}
  \includegraphics[width=1.0\linewidth]{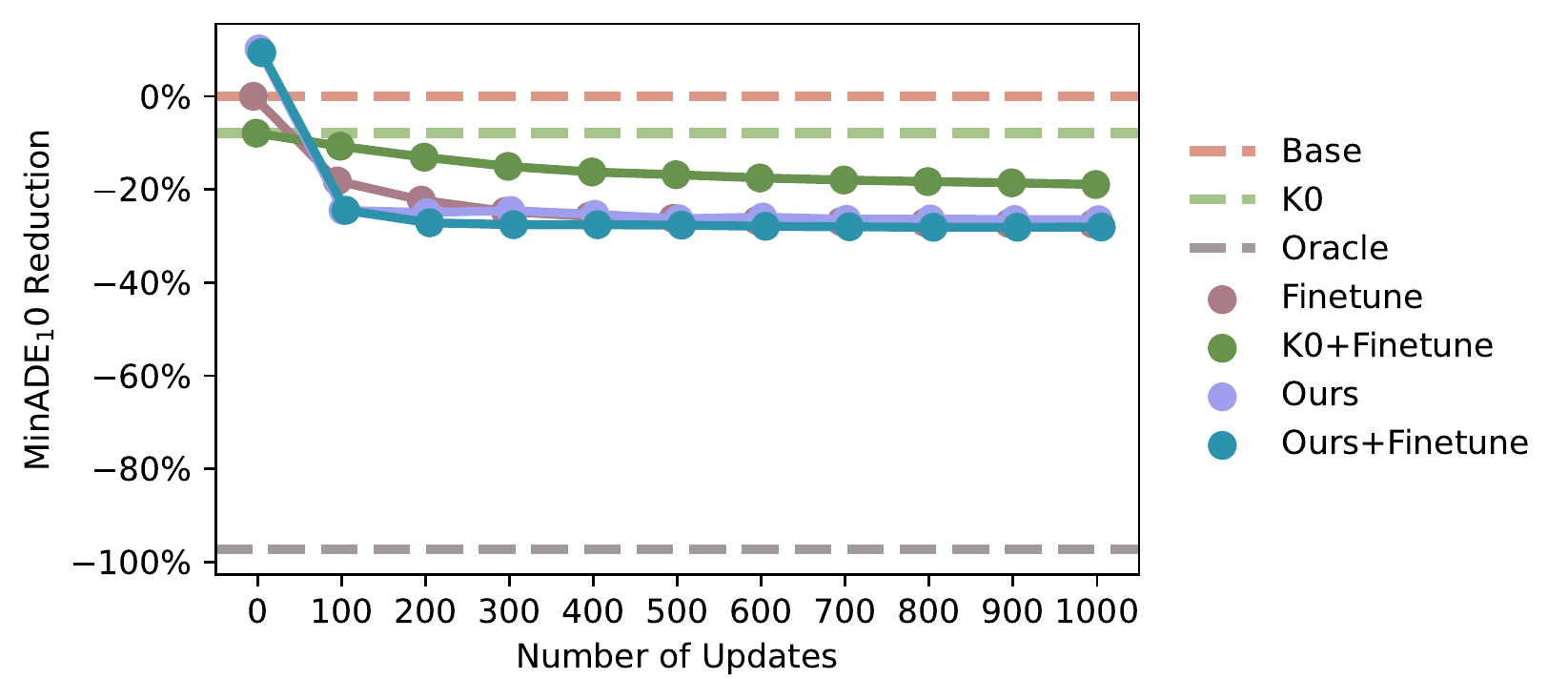}
  \caption{[nuScenes $\rightarrow$ Lyft] Our method's gradual improvement with increasing amounts of data in the offline problem setting can be seen in other metrics, confirming the result in \cref{fig:lyft_offline}. Lower is better.}
  \label{fig:supp_lyft_offline}
\end{figure}

\begin{figure*}[t]
  \centering
  \includegraphics[height=0.18\linewidth]{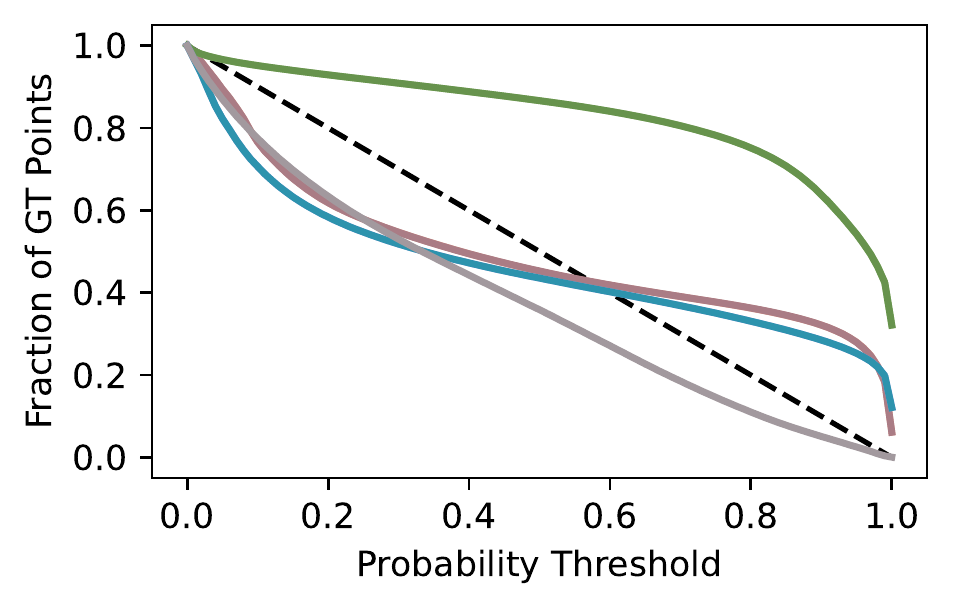}
  \includegraphics[height=0.18\linewidth]{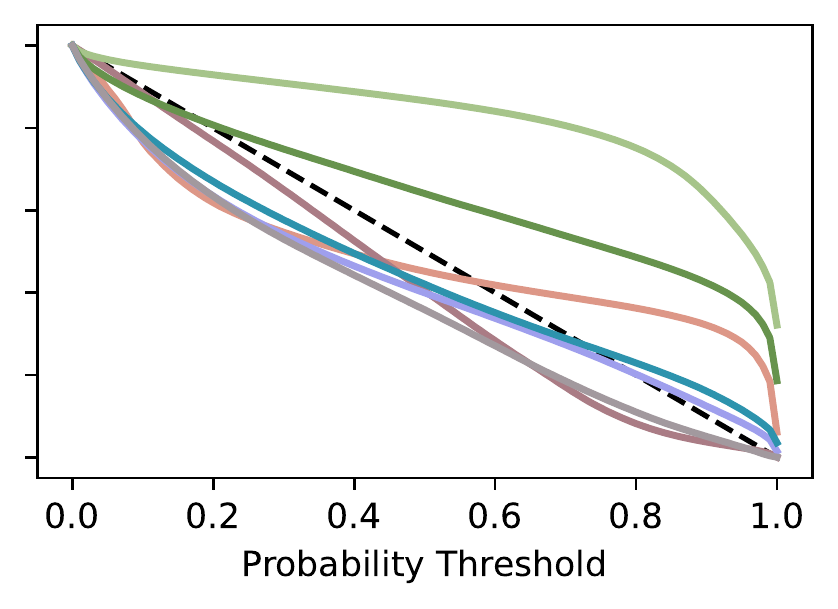}
  \includegraphics[height=0.18\linewidth]{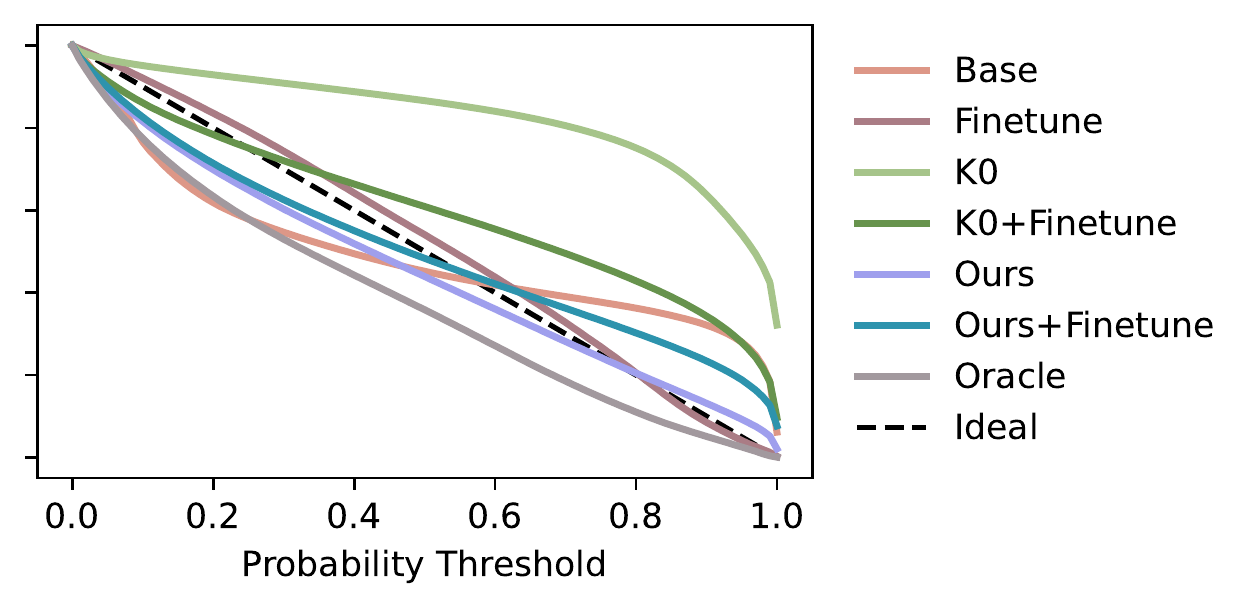}
  \caption{[nuScenes $\rightarrow$ Lyft] Detailed calibration plots showing the fraction of GT future positions lying within specified probability thresholds after observing 0, 500, and 1000 data samples in the offline setting. Our method fast approaches the ideal line (with Ours+Finetune doing so even faster), and stays better calibrated than only performing whole-model finetuning (Finetune). The area below the ideal line signifies underconfidence (e.g., the oracle model). Accordingly, the area above the ideal line signifies overconfidence (especially visible with $K_0$). Computing the area between each line to the ideal line yields the ECE values after 0, 500, and 1000 updates in \cref{fig:lyft_offline_ece}.}
  \label{fig:supp_lyft_offline_calib}
\end{figure*}

\begin{figure}[t]
  \centering
  \includegraphics[width=1.0\linewidth]{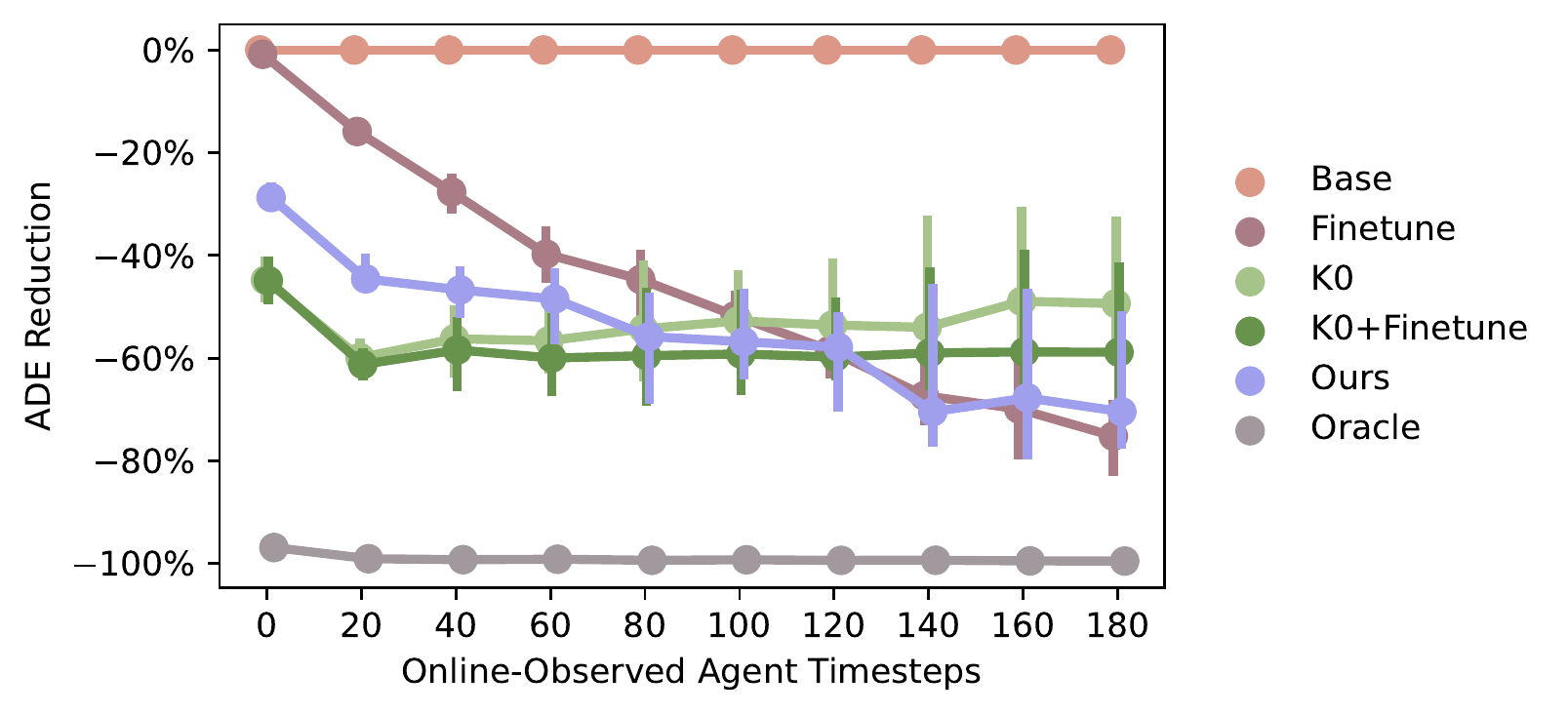}
  \includegraphics[width=1.0\linewidth]{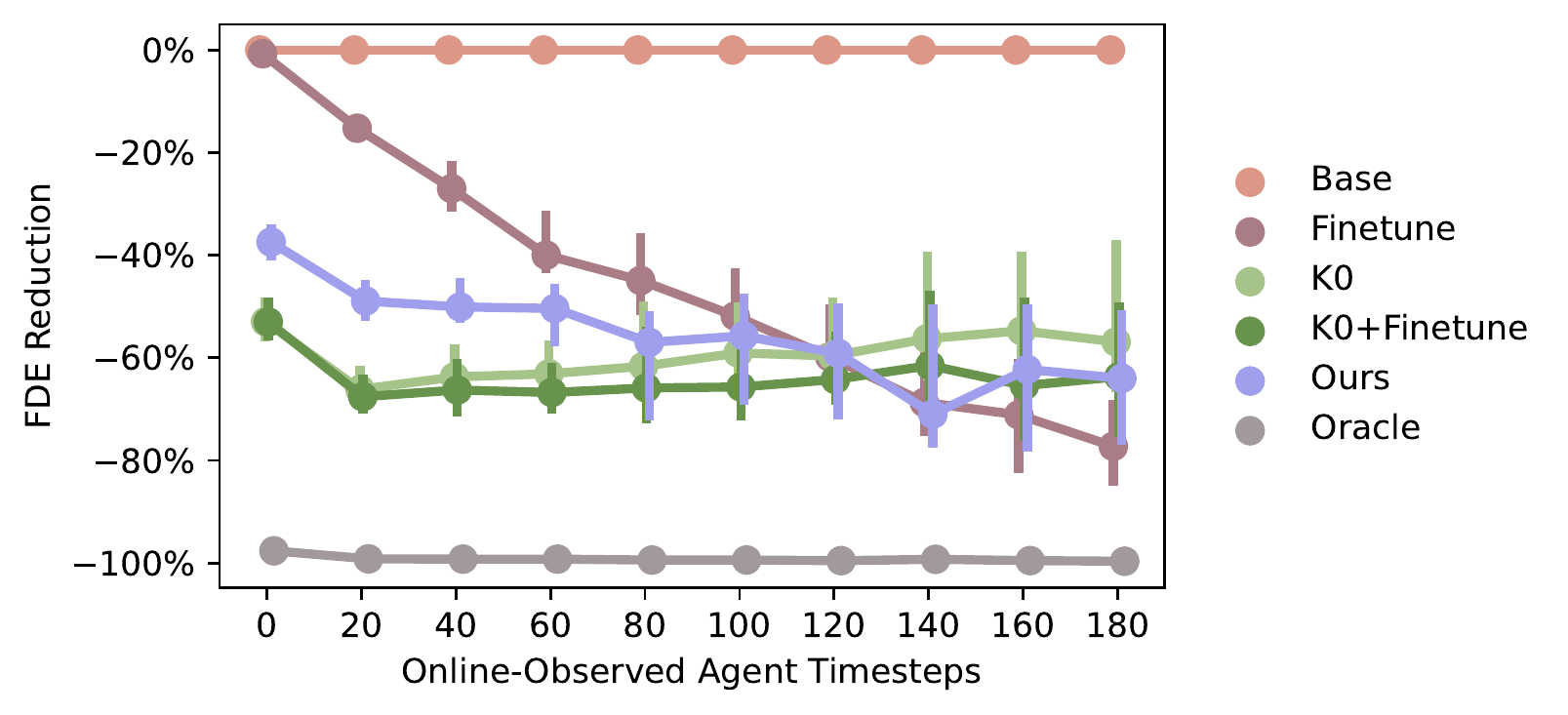}
  \includegraphics[width=1.0\linewidth]{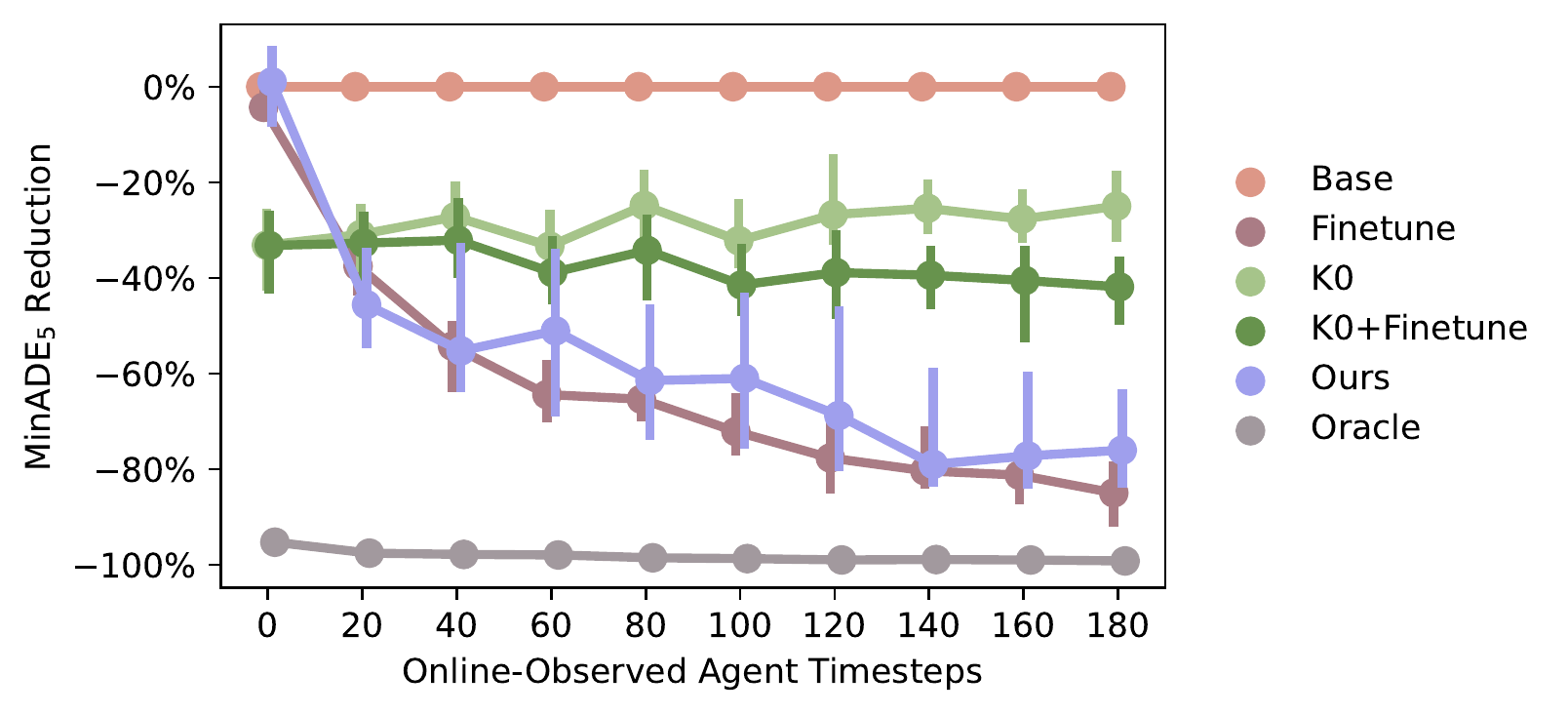}
  \includegraphics[width=1.0\linewidth]{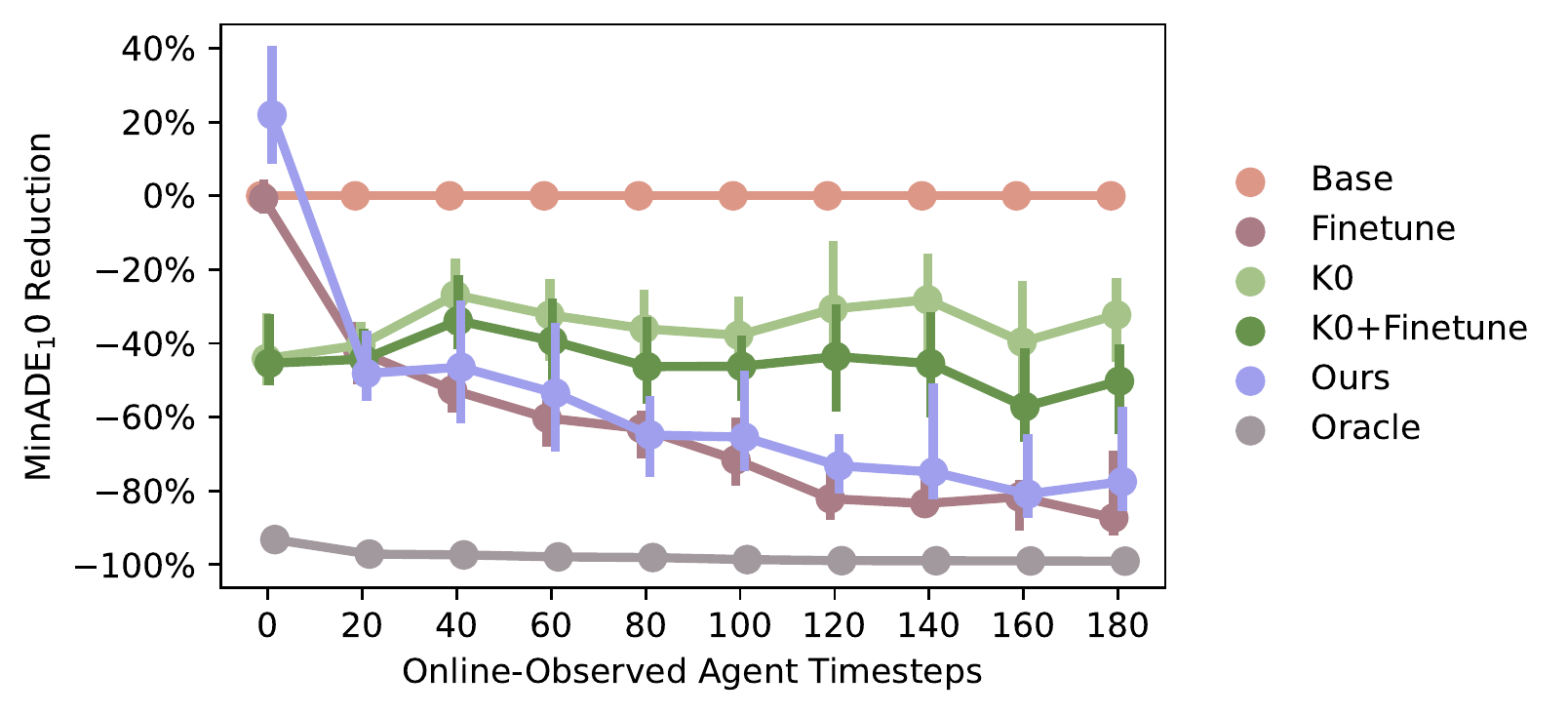}
  \caption{[nuScenes $\rightarrow$ Lyft] Evaluating our approach on additional metrics matches the result in \cref{fig:lyft_online}; our approach rapidly improves online as it observes more data. Error bars are 95\% CIs, lower is better.}
  \label{fig:supp_lyft_online}
\end{figure}

\subsection{Computational Complexity}\label{sec:supp_computation}

There are several considerations that are important for achieving efficient performance in the model. First, in practice, we fix a set of parameters for each output dimension, which are adapted independently. This corresponds to choosing a diagonal noise and parameter covariance, fixing parameter dynamics ($A$) diagonal (in practice we simply choose this to be identity), and fixing independent priors for parameters of each output dimension. Parameterizing each output dimension independently substantially reduces computational complexity, as discussed in~\cite{LewEtAl2022}.

Furthermore, we may implement the model such that complexity is at most quadratic in the parameter dimension. First, $A$ must be chosen appropriately; identity is sufficient but other representations are possible. In addition, the multiplication of $K_{t+1} \Feat_{t+1} S_{t+1|t}$ should be done with the last two terms first, before multiplying by the gain matrix $K_{t+1}$. This enables better representational capacity for adaptation by enabling choice of a larger number of features. 

Finally, we note that prediction in the model is sequential in time due to the unrolling of the recurrent model. Thus, at run time, parallelizing over a large number of predictions is straightforward. Indeed, for hardware such as GPUs or other specialized cores for neural networks, parallelized matrix multiplications are extremely efficient, and thus many samples can be parallelized across.

\subsection{Additional Pedestrian Evaluations}\label{sec:supp_peds}

\cref{tab:supp_peds_d2d} shows the same analyses as in \cref{tab:peds_d2d}, but with the FDE evaluation metric.
\cref{fig:supp_peds_online} shows the same analyses as \cref{fig:peds_online}, but with the FDE and NLL evaluation metrics.
Lastly, \cref{fig:supp_peds_offline_calib} shows detailed calibration plots which underlie the ECE computations in \cref{fig:peds_offline_ece}.
Broadly, the results of these additional evaluations confirm and reinforce the performance of our approach over other baselines and ablations.

\subsection{Additional Autonomous Driving Evaluations}\label{sec:supp_lyft}

\cref{fig:supp_lyft_offline} shows the same analyses as \cref{fig:lyft_offline}, but on the other four evaluation metrics.
\cref{fig:supp_lyft_offline_calib} shows detailed calibration plots which underlie the ECE computations in \cref{fig:lyft_offline_ece}.
Finally, \cref{fig:supp_lyft_online} shows the same analyses as in \cref{fig:lyft_online}, but on the other four evaluation metrics.
Overall, the results of these additional evaluations confirm and reinforce the performance of our approach over other baselines and ablations.

\end{document}